\documentclass[preprint,12pt]{elsarticle}

\usepackage{graphicx,}
\usepackage{bm}
\usepackage{subfigure}
\usepackage{amssymb}
\usepackage{balance}
\usepackage{algorithm}
\usepackage{algorithmic}
\usepackage{booktabs}
\usepackage{multirow}
\usepackage{colortbl}
\usepackage{url}

\journal{arXiv}

\begin{document}

\begin{frontmatter}



\title{Intelligent gradient amplification for deep neural networks.}


\author[gsutrnd,gsucs]{Sunitha Basodi\corref{cor1}}
\author[gsutrnd]{Krishna Pusuluri}
\author[gsucs]{Xueli Xiao}
\author[cnun]{Yi Pan}

\cortext[cor1]{sbasodi1@gsu.edu}
            
\affiliation[gsutrnd]{organization={Tri-institutional Center for Translational Research in Neuroimaging and Data Science (TReNDS), Georgia State University, Georgia Institute of Technology, Emory University},
            addressline={55 Park Place}, 
            city={Atlanta},
            postcode={30303}, 
            state={Georgia},
            country={USA}}
            
\affiliation[gsucs]{organization={Department of Computer Science, Georgia State University},
            addressline={25 Park Place NE}, 
            city={Atlanta},
            postcode={30303}, 
            state={Georgia},
            country={USA}}
               
\affiliation[cnun]{organization={Faculty of Computer Science and Control Engineering, Shenzhen Institute of Advanced Technology, Chinese Academy of Sciences},
            addressline={Room D513, 1068 Xueyuan Avenue,Shenzhen University Town}, 
            city={Shenzhen},
            postcode={518055}, 
            country={China}}

\begin{abstract}
Deep learning models offer superior performance compared to other machine learning techniques for a variety of tasks and domains, but pose their own challenges. In particular, deep learning models require larger training times as the depth of a model increases, and suffer from vanishing gradients. Several solutions address these problems independently, but there have been minimal efforts to identify an integrated solution that improves the performance of a model by addressing vanishing gradients, as well as accelerates the training process to achieve higher performance at larger learning rates. In this work, we intelligently determine which layers of a deep learning model to apply gradient amplification to, using a formulated approach that analyzes gradient fluctuations of layers during training. Detailed experiments are performed for simpler and deeper neural networks using two different intelligent measures and two different thresholds that determine the amplification layers, and a training strategy where gradients are amplified only during certain epochs. Results show that our amplification offers better performance compared to the original models, and achieves accuracy improvement of $\sim$2.5\% on CIFAR-10 and $\sim$4.5\% on CIFAR-100 datasets, even when the models are trained with higher learning rates.
\end{abstract}





\end{frontmatter}


\section{Introduction}

Deep learning models have produced results comparable or sometimes superior to human experts in many interdisciplinary applications\cite{ gopalakrishnan2017deep, wang2019deep, yan2018multi,2022liuminresearch, basodi2020data}.  Performance of these models generally improves with increasing depth of the network \cite{huang2016deep}, but some challenges arise such as vanishing gradients and high training time even on parallel computational resources \cite{huang2016deep}. There are many architectures \cite{haber2017stable} that work well for a wide range of applications but even the simplest models are computationally intensive. As there are millions of model parameters for each set of hyper parameters, deep neural networks are slow to train, sometimes ranging from several days or weeks depending on the model architecture, dataset size, and hardware resources. One way to improve the training time is to do this in parallel using Graphical Processing Units(GPUs), or Tensor Processing Units(TPUs), or on parallel distributed systems. The other way to further improve is to train the model with higher learning rates or large batch sizes. Employing larger learning rates can speedup the training process by quickly converging to local optima, but they can often miss the global optima, resulting in sub-optimal solutions or sometimes non-convergence \cite{art2_learningrate_vs_time}. 
Lower learning rates converge better to optima, but have larger training times. In general, a combination of higher and lower learning rates  are employed, using scheduling schemes or algorithms which adaptively reduce the learning rates over epochs. There have been multiple efforts to analyze and modify gradients and learning rates dynamically during the training process, but there is no detailed analysis on the impact of the modification factor on model performance.
Vanishing gradients \cite{hochreiter2001gradient, goh2017deep, hanin2018neural} is another problem that can occur while training deep neural networks. Some approaches help to mitigate this issue, such as weight initialization followed by fine-tuning with backpropagation \cite{schmidhuber1992learning}, using Rectified Linear Unit (ReLU) activation function \cite{nair2010rectified, glorot2011deep}, or applying batch normalization (BN) \cite{ioffe2015batch}. In addition, advancements in hardware capablities of GPUs such as double precision help overcome this issue to some extent while training deep neural networks. 

Though there have been efforts to address the above mentioned problems independently, there have been minimal efforts to identify an integrated solution to improve the performance of the model by addressing both vanishing gradients and accelerating the training process by achieving higher performance at larger learning rates. To improve the training process of existing deep learning models, we propose an intelligent gradient amplification approach, along with a training strategy that analyzes the net gradient change of a layer in an epoch. While performing gradient amplification, gradients are dynamically increased for some layers during back propagation using a training strategy where amplification is performed for a few epochs, while the model is trained without gradient amplification for the remaining epochs. Neural networks trained using our method have improved testing and training accuracies, even at higher learning rates (thereby reducing the overall training time).


In this work, we extend the gradient amplification method proposed in \cite{basodi2020gradient} using a formulated approach by analyzing the fluctuations of gradients within and across layers during training, and intelligently identifying the layers to perform gradient amplification. Our contributions include the following:

\begin{itemize}
	\item[--] We propose two measures to compute the effective gradient update direction of each layer, which are independently analyzed by performing normalization.
	\item[--] We suggest two thresholding approaches to intelligently identify the layers to amplify, one when the actual normalized measure of a layer crosses the threshold and the other when the absolute normalized measure is beyond the threshold.
	\item[--] We study the impact of thresholds by analyzing a wide range of values and identify threshold values that could generally work for a deep learning model.
	\item[--] We perform detailed experiments using two intelligent measures with two thresholding approaches applying a training strategy where gradients are amplified during certain epochs and the model is trained in the remaining epochs without amplification.

\end{itemize}

The remainder of this paper is organized as follows. Related works are briefly described in Section \ref{sec:related_work}. Our amplification method with proposed measures and thresholding approaches is presented in Section \ref{sec:problem_formulation}. Experimental setup, results, and their comparisons are discussed in Section \ref{sec:intel_amp_experiments_form1} and \ref{sec:results}, followed by conclusions in Section \ref{sec:conclusion}.

\section{Related Work }\label{sec:related_work}

Although deep learning models are robust and acheive better performance, there are several areas where these models could be improved further. Designing new architectures, automatic tuning of network hyperparameters, improving training time of the models, designing efficient functions (activation, kernal and pooling are some of such challenges\cite{xiao2020efficient, rawat2017deep, szegedy2013deep}. In this section, we briefly discuss the existing approaches to address vanishing gradient problem, reduce the training time of deep learning models, and study the impact of learning rates.

\subsection{Vanishing gradients} 
The problem of vanishing gradients \cite{hochreiter2001gradient, goh2017deep, hanin2018neural} occurs in gradient-based learning methods while training models with backpropagation.  As the network weights get updated based on the gradients during back-propagation, lower gradient values cause the network to learn slowly and the value of these gradients reduces further while propagating from end layers to initial layers of the network. When gradients are close to 0, a model cannot learn while training and the weight updates do not have any effect on its performance. Though there is no direct solution for this issue, several methods suggested in the literature \cite{schmidhuber2015deep} help to mitigate the issue. One of the early proposed method \cite{schmidhuber1992learning} involves a two step weight update rule, where, network weights are updated based on unsupervised learning methods and then fine-tuned using supervised learning with back-propagation.  In recent years, with the introduction of ReLU activation function \cite{nair2010rectified, glorot2011deep}, batch normalization(BN)\cite{ioffe2015batch} and Resnet networks\cite{he2016deep}, this problem reduced further. When a model has ReLU activation function\cite{nair2010rectified, glorot2011deep}, only positive inputs get propaged in the forward pass, it is observed from experiments that the gradients in the backward pass do not diminish and prevents vanishing gradients to some extent. The other approach is to use BN\cite{ioffe2015batch} layers in a model. These layers normalize the inputs to reduce its variance during the forward pass and therefore help in regulating the gradients in the backward pass during training. One can use both ReLU activation layer and BN layer in a network, improving the performance of the model even further while training. Resnet networks use these combination of layers in addition to other layers and also have residual connection in the network, connecting some initial residual blocks(or layers) to the later residual blocks(or layers). Since the gradients also get directly propaged with these connections to the initial layers in addition to the network layers, the problem of vanishing gradients reduces even futher. 
In addition to these approaches, with the constant advancement in the hardware and their increased computation abilities, the problem has further reduced. In out method, as we increase the gradients of some of the layers of the network during backpropagation, it helps to aid the problem of vanishing gradients.

\subsection{Learning rates} 

Learning rate is one of the critical hyperparameters that directly influences the performance of the deep learning models. Models can be trained faster with larger learning rates, but can lead to sub-optimal solutions. Using lower learning rates converges the models better with optimal solutions \cite{art2_learningrate_vs_time} but has long training times. There are many approaches designed to achieve better performance using a combination of these learning rates.  Learning rate scheduler is one such ways, where the training begins with higher learning rates which is lowered as the training continues \cite{darken1992learning}.  Lowering of learning rates in a scheduler can be designed in many ways \cite{art1_learningrate} , such as, assigning learning rates to epochs, gradually decaying the learning rate based on the current learning rate, current epoch and total number of epochs(time-based decay); reducing the learning rate in a step-wise manner after a certain number of epochs(step decay); and exponentially decaying the learning rate based on the initial learning rate and the current epoch(exponential decay). There are some other approaches suggested, such as, normalizing model weights while training with stochastic gradient descent to speed up the performance \cite{salimans2016weight}. Authors of \cite{smith2017don} increase batch size without decaying learning rates and observe that the models achieve similar test performances. Since this uses larger batch sizes, it increases parallelism and has fewer parameter updates thereby reducing the overall training times.

In general, while training a model with learning rate scheduler, higher learning rates used in the beginning for a few epochs, followed by lower of learning rates for the next few epochs; and this process is repeated until the desired optima or model performance is achieved. Some optimization algorithms automatically determine the learning rates with the epochs dynamically without the requirement of manual intervention. However, these methods also have some fallbacks and do not always converge to optimal solutions. One way to improve the training speed is to develop methods to achieve optimal model parameters at larger learning rates.

\subsubsection{Adaptive learning rates for layers/neurons/parameters}
Another approach to overcome identifying learning rate hyperparameters without need of any scheduling is by modifying learning rate dynamically based on the performance of the optimization algorithm. A few of such methods include Adagrad\cite{duchi2011adaptive},  Adadelta\cite{zeiler2012adadelta}, RMSprop\cite{graves2013generating} and Adam \cite{kingma2014adam}. There have also been several efforts to improve models with adaptive learning rates \cite{yu2002backpropagation, dauphin2015equilibrated, liu2019variance,luo2019adaptive, gupta2019finite}. Ede et al. \cite{ede2020adaptive} control the gradients from being propagated when the expected loss is over a defined boundaries. This causes the learning rates to be dynamically addjuste during training. Paper \cite{daniel2016learning} designs a controller to automatically manage learning rate by identifying informative features. Authors of \cite{xu2019learning} identify learning rates using reinforcement learning based approach by analyzing the training history. Experiments are performed on CIFAR-10 AND FMNIST database have improved performance emphasizing its advantages. You et al. \cite{you2017large} adaptively scale the learning rates of layers to improve the performance of models trained in large batches in parallel and perform experiments on AlexNet\cite{krizhevsky2017imagenet}. Paper \cite{zhang2018train} proposes a layer-wise adaptive learning rate computation by using layer weight dependent matching factor which is computed dynamically during training based on the layer type. Authors demonstrate the advantage of their method with mathematical equations and experimental results.

\subsection{Analysis of gradients}
Gradients provide vital information on various aspects such as training progression, weight fluctuations, model convergence and so on. This information can be used to address vanishing/exploding gradients, accelerate the training process or dynamically modify learning rates while training the model. Some of the adaptive learning rate algorithms \cite{duchi2011adaptive}\cite{zeiler2012adadelta} mentioned above use the gradients of a few iterations to identify a suitable learning rate on the current iteration. Zhang et al. \cite{zhang2018train} scale the gradients of a layer based on a matching factor which is computed during training based on the weights and type of the layer.

\section{ Proposed Gradient Amplification Strategies }\label{sec:problem_formulation}

Gradient amplification with random selection of layers proposed in \cite{basodi2020gradient} has overhead of identifying the ratio of layers, the type of layers and the combination of those types of layers. With the increase in number of layers and also with different types of layers, the network becomes deeper and it is challenging to determine the best amplified models of all the combinations. In this section, we aim to formulate a way to automatically determine the layers which need to be amplified.

\subsection{Effect of gradient amplification on learning rate}
Here, we emphasize the relationship between learning rate and its effects on gradients while performing amplification.  The general weight update formula during training of neural networks is shown below: 

\begin{equation}
\bm{W}_{t+1}= \bm{W}_t + \eta \nabla J(\bm{W}_{t+1}) 
\label{eq:grad_rule}
\end{equation}

where, $\bm{W}_t$ represents the weights of a network in the current iteration $t$, $\eta$ is the learning rate and $ \nabla J(\bm{W}_{t+1})$ corresponds to the gradients of the weights computed as the derivative of error of cost function with respect to weights.

After performing gradient amplification, $ \nabla J(\bm{W}_{t+1})$ gets modified depending on the amplification factor. Let us denote the amplified gradient as
\begin{equation}
 \nabla J_{amp} =  amp * \nabla J(\bm{W}_{t+1}) 
 \label{eq:amp_effect}
\end{equation}

Therefore the weight update formula after gradient amplification can be written as:

\begin{eqnarray}
&\bm{W}_{t+1}= \bm{W}_t + \eta \nabla J_{amp}(\bm{W}_{t+1}) \nonumber \label{eq:derivation_grad_amp_lr} \\
&\bm{W}_{t+1}= \bm{W}_t + \eta * amp * \nabla J(\bm{W}_{t+1})  \\
&\bm{W}_{t+1}= \bm{W}_t + (\eta * amp) * \nabla J(\bm{W}_{t+1}) \nonumber \\
&\bm{W}_{t+1}= \bm{W}_t + \eta_{amp} * \nabla J(\bm{W}_{t+1}),  \; where \; \eta_{amp}=\eta * amp \nonumber 
\end{eqnarray}

From the above analysis, one can conclude that amplifying gradients is equivalent to increasing learning rates. To determine the layers that need to be amplified, one approach is to identify the layers that learn actively during the training process.  Authors of \cite{xiao2019fast} propose a way to speedup the training process of deep learning models using layer freezing approach by determining the fluctuations of gradients in layers. Similar approach can be employed to determine the layers that are actively learning. As performing amplification is equivalent to increasing step size of weight updates, learning rate (or step size) can be increased when the current weights of the neurons are relatively far from optima and their gradients are all moving in the same direction to converge to optima. In addition to determining the layers that are actively learning, it is also important that the gradients of the neurons in the layers are all moving in the same direction for the amplification to be meaningful. This can be formulated by analyzing the gradients of the neurons in a layer across iterations. One simple way to perform such an identification is to compute the sum of the gradient values over iterations for all neurons in a layer. 

\subsection{Layer gradient directionality ratio measure, $G$}

The effective direction of gradient change of the neurons in a layer $l$  can be determined using the ratio of the sum of the gradients across different iterations to its absolute gradient sum in an epoch.  Here, $m$ and $n$ correspond to the number of iterations in an epoch and the number of neurons(or weights) in a layer respectively.


\begin{equation}
\label{form:formula_1_definition}
G_{l} = \cases{0, \; when\; \sum_{i}^{n}\sum_{j}^{m}| g_{lij}|=0 \\
 \frac{ \sum_{i}^{n}|\sum_{j}^{m} g_{lij}|}{ \sum_{i}^{n}\sum_{j}^{m}| g_{lij}|}, \; otherwise \\}
\end{equation}

The above formula determines how the weights in a layer are modified. When all the weight updates of the neurons occur in the same direction across all iterations, its value is 1. When either the model reaches optimal solution (where the gradient for every neuron becomes zero) or half of the neuron weights move in the opposite direction to the other half with same magnitude (ideal case) then the value becomes 0. Otherwise, it lies in between 0 and 1. Values close to 1 signify that most of the weights are changing in the same direction and vice versa.

\begin{equation}
0 \leqslant G_{l} \leqslant 1
\label{form:gradient_change_range_form1}
\end{equation}

\subsection{Normalized layer gradient directionality ratio measure, $\hat{G}$}
After computing the \textit{layer gradient directionality ratio ($G_{l}$) }for each layer in an epoch, it is observed that most of these values are in same range. To identify the most significant layers close to 1, \textit{layer gradient directionality ratio ($G_{l}$)} of all the layers are converted to a normal distribution using :


\begin{equation}
 \hat{G}= \frac{G - \overline{\rm G}  } { \sigma_{G}  }
 \label{form:formula_1_normalized}
\end{equation}

These normalized values signify how far they lie from the mean value. If the normalized value of a layer is close to 0, then it lies close to the mean value, otherwise, its value signifies the magnitude of standard deviation(s) it is away from the mean value. Larger the positive(or negative) value farther it is from the mean. Since the range of $G{l}$ is $[0, 1]$, when the normalized value is a large positive value (signifying farther from the mean on the right side), it can be considered closer to the 1 (in equation \ref{form:gradient_change_range_form1}), where as large negative values can be considered close to 0 (in equation \ref{form:gradient_change_range_form1}).

\subsection{Improved Layer gradient directionality ratio measure, ${G'}$}
The gradients of neurons of a layer lie in a similar range and the measure $G$, which has the sum of all the gradients of neurons across all the iterations, gives more importance to the gradients values that occur frequently during the training. If equal importance is given to the all the neurons, then even if some of the ratio measures are close to 1, then the layer has relatively higher ratio value (closer to 1). In contrast, when the gradient changes for each neuron across multiple iterations are computed individually and are  used for computing effective directionality of layer, all neurons then contribute equally towards the gradient direction change. With this modification, even if some of the neuron gradients in a layer are moving in the same direction, the layer still has an higher chance of being identified.

The effective direction of gradient change of the neurons in a layer $l$ is therefore determined using the ratio of the sum of the gradients across different iterations to its absolute gradient sum for each neuron and then taking the mean of these ratio values.   Here, $m$ and $n$ correspond to the number of iterations in the epoch and number of neurons(or weights) in a layer respectively for an epoch.


\begin{equation}
G'_{l} =  \cases{
0, \; when \; \sum_{j}^{m}| g_{lij}|=0 \\
\frac{1}{n}\sum_{i}^{n}  \frac{ |\sum_{j}^{m} g_{ij}|}{ \sum_{j}^{m}| g_{ij}|} ,  \; otherwise \\
}
\label{form:formula_2_definition}
\end{equation}

The above formula also determines how the weights in a layer are modified with equal importance given to all the neurons in a layer. When all the weight updates of the neurons occur in the same direction across all iterations, its value is 1. With the modified formula, when some of the neurons are close to 1, it have higher chances of being amplified.

\begin{equation}
0 \leqslant G'_{l} \leqslant 1
\label{form:gradient_change_range_form2}
\end{equation}

\subsection{Normalized layer gradient directionality ratio measure, $\hat{G'}$}
After computing the\textit{ layer gradient directionality ratio ($G'_{l}$)} for each layer in an epoch, in order to identify the most significant layers close to 1, \textit{layer gradient directionality ratio ($G'_{l}$)} of all the layers are converted to a normal distribution using :


\begin{equation}
 \hat{G'}= \frac{G' - \overline{\rm G'}  } { \sigma_{G'}  }
 \label{form:formula_2_normalized}
\end{equation}

As previously mentioned, these normalized values signify how far they lie from the mean value. 
As $G'_{l}$ ranges between $0$ and $1$,  large positive normalized values (signifying farther from the mean on the right side), can be considered close to the 1 (see equation \ref{form:gradient_change_range_form2}), where as large negative normalized values can considered close to 0 (see equation  \ref{form:gradient_change_range_form2}).

\subsection{Determining amplification layers using $G$,$G'$ measures}
We consider the following cases to perform amplification based on thresholds. Here the formula are shown for measure $G$ and the same cases can be applied for $G'$ (by replacing $G$ with $G'$).

\paragraph{Case-1: Amplify only one side ($G_{l}$ close to 1)}
When $(G_{l})$ values are close to 1, most of the weights in the layer are modified in the same direction. This suggests that the neurons are all moving down(or up) the slope to approach optima, learning rate can be increased. We amplify the gradients of the layer when its normalized value exceeds a threshold value,

\[
	if (\hat{G} > threshold ) \; :\; amplify \; layer
\]

\paragraph{Case-2: Amplify both sides ($G_{l}$ close to 0 or 1)}
As mentioned earlier, the $layer\_gradient\_directionality\_ratio$ $(G_{l})$ values close to 0 either mean they are close to optima or a plateau surface. If the weights are close to optima, adding small noise to the gradients will cause the weights to converge eventually. Otherwise, it would make the weights to cross the plateau surface and thereby improving the training process. With this assumption, we propose to amplify the gradients of a layer when the $layer\_gradient\_directionality\_ratio$ $(G_{l})$ values are close to either end (i.e., 0 or 1), we amplify whenever the absolute normalized value crosses threshold value.

\[
	if (|\hat{G}| > threshold ) \; :\; amplify \; layer
\]

Here is the overview of the function to determine amplification layers

\begin{algorithm}
\caption{Determination of $amp$ layers using Formula-1}
\label{alg:find_amp_formula1}
\begin{algorithmic}
\renewcommand{\algorithmicrequire}{\textbf{Input:}}

\REQUIRE $M$, $threshold$, $g_{lij}$, $case$


\renewcommand{\algorithmicrequire}{\textbf{Function:}}
  \REQUIRE \textsc{ GetGradientAmpLayers({$M$, $threshold$, $g_{lij}$, $case$}) } \label{alg:formula_1_func}

  \STATE Compute $layer\_gradient\_directionality\_ratio (G_{l})$ as in equation (\ref{form:formula_1_definition})
  \STATE Compute normalized values $\hat{G}$ as in equation (\ref{form:formula_1_normalized})
  \IF {$case == 1$}
      \IF {$\hat{G_l} > {threshold}$}
        \STATE add layer $l$ to $amp$
      \ENDIF
  \ELSE
         \IF {$case == 2$}
           \IF {$mod(\hat{G_l}) > threshold$}
              \STATE add layer $l$ to $amp$
           \ENDIF
         \ENDIF
\ENDIF
 \RETURN $amp$
 \renewcommand{\algorithmicrequire}{\textbf{EndFunction}}
 \REQUIRE $ $

\end{algorithmic}
\end{algorithm}

\section{Experiments}\label{sec:intel_amp_experiments_form1}

\begin{figure}
\includegraphics[width=0.5\textwidth]{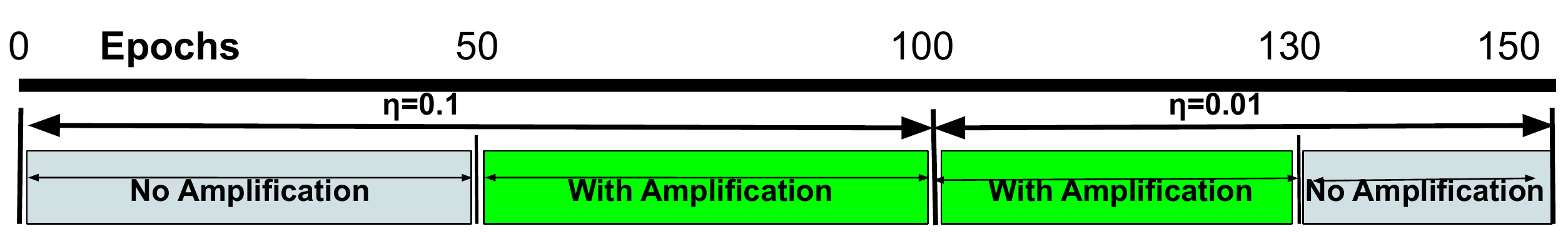}
\caption{ Experimental setup and training strategy for all the models, showing the number of epochs and the corresponding learning rates($\eta$).}
\label{fig:GradAmp_Epochs_3}
\end{figure}

Experiments are performed on CIFAR-10, CIFAR-100 datasets with the similar setup to the experiments performed as in paper \cite{basodi2020gradient}. We primarily perform thorough experiments for VGG-19, Resnet-18 and Resnet-34 models. These models are trained for 150 epochs, where the learning rate of the first 100 epochs is 0.1 and the next 50 epochs is 0.01. 
We perform experiments using the training strategy $params_1=[(50, 0.1, 0, 1), (100, 0.1, is\_amp, 2), (130,  \linebreak0.01, is\_amp, 2), (150, 0.01, 0, 1)]$, as shown in Fig. \ref{fig:GradAmp_Epochs_3}, where $is\_amp$ represents a non-zero value when amplification is performed or 0 otherwise. The values in each element in the $params$ list represent the end epoch, learning rate, is\textunderscore amplification\textunderscore performed(non-zero) and gradient amplification factor respectively. For instance, $(50, 0.1, 0, 1)$ means that the model is trained with learning rate $0.1$ until we reach $50$ epochs, during which no layers are selected for gradient amplification and the amplification factor is $1$. In our training strategy, where $params_1 = [(50, 0.1, 0, 1), (100, 0.1, is\_amp, 2), (130, 0.01, \linebreak is\_amp, 2), (150, 0.01, 0, 1)]$, no amplification is performed for the first 50 epochs and gradients in the $51^{st}$ epoch are analyzed to determine the layers to perform amplification until $100^{th}$ epoch. At epoch 101, learning rate is reduced to 0.01 and gradients are analyzed again to determine next set of amplification layers, where the selected layers are amplified for the next 29 epochs and the last 20 epochs are trained without amplification.  Experiments are performed using varying thresholds from 0.7 to 2.5 with the step size of 0.1. Based on the analysis of these results, we also run experiments on even deeper resnet-50 and resnet-101 models.

 \begin{figure*}
  \centering
      {\footnotesize {Frequency of amplification analyzed on CIFAR-10 dataset using measure $G$  case-2}}\par\medskip

   \vspace*{-0.2cm}
    \subfigure[$VGG-19$\label{fig:Formula_1_case_3_steps_params_1_VGG_19}]{\includegraphics[width=0.33\textwidth]{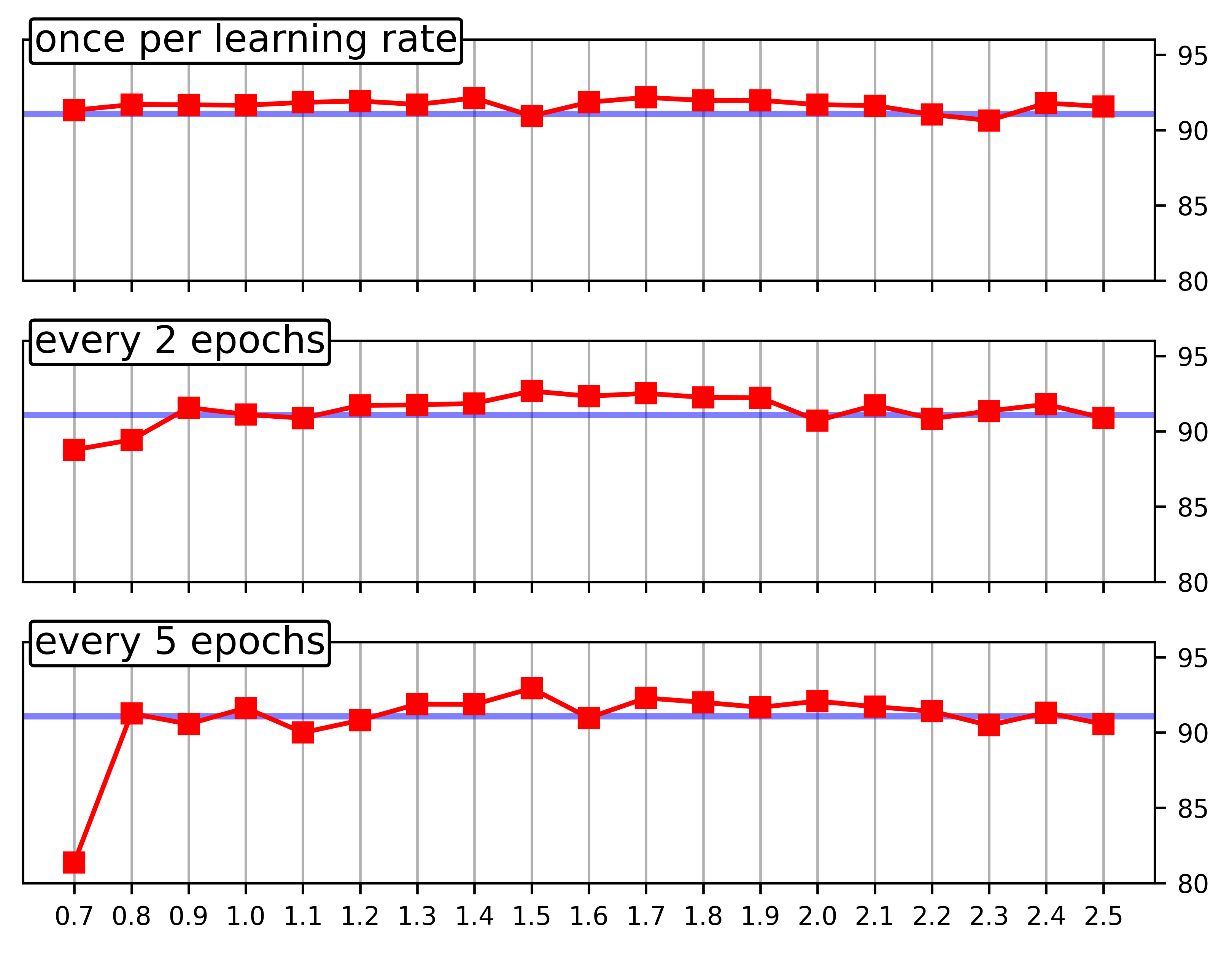}}
    \subfigure[$Resnet-18$\label{fig:Formula_1_case_3_steps_params_1_Resnet_18}]{\includegraphics[width=0.33\textwidth]{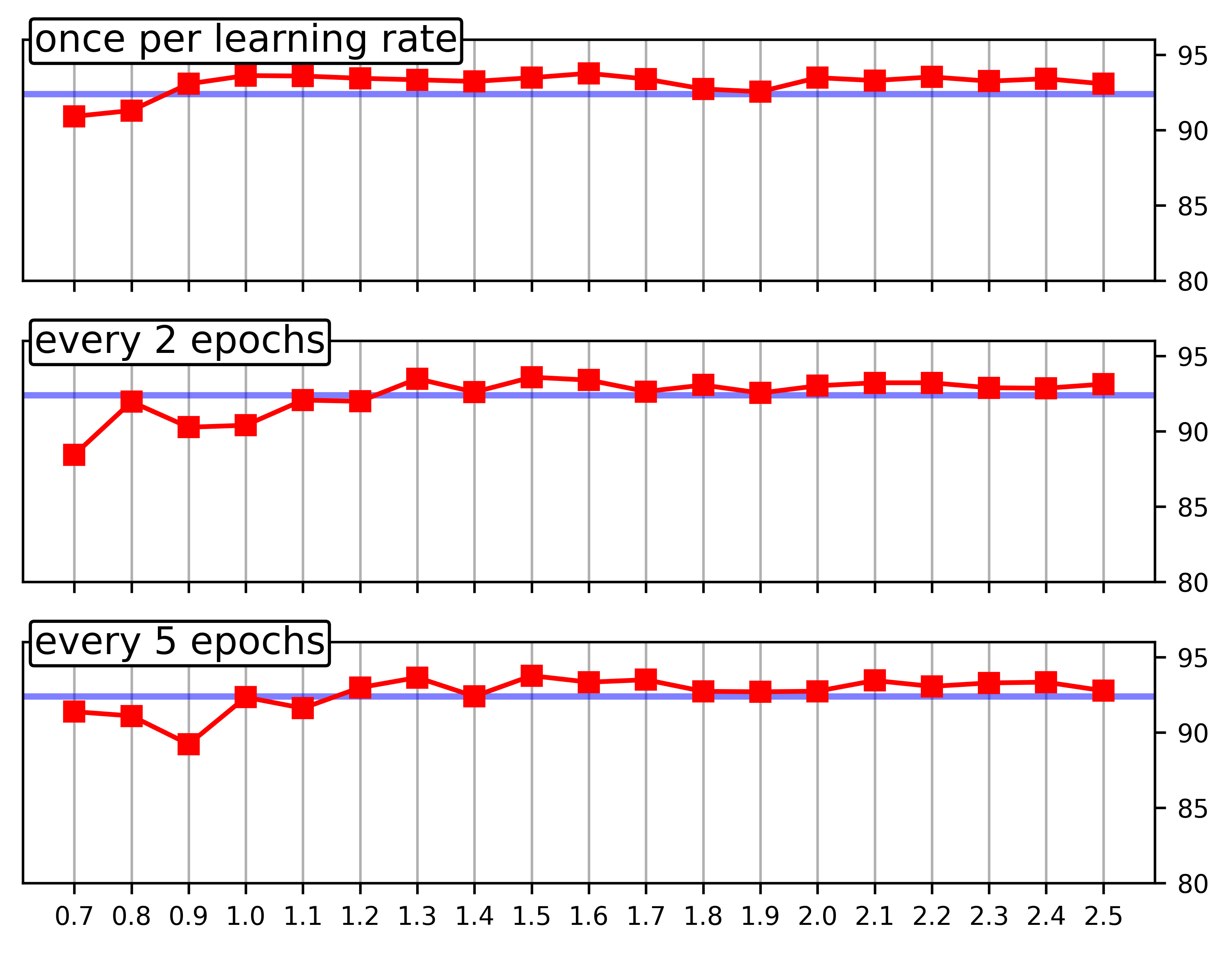}}
    \subfigure[$Resnet-34$\label{fig:Formula_1_case_3_steps_params_1_Resnet_34}]{\includegraphics[width=0.33\textwidth]{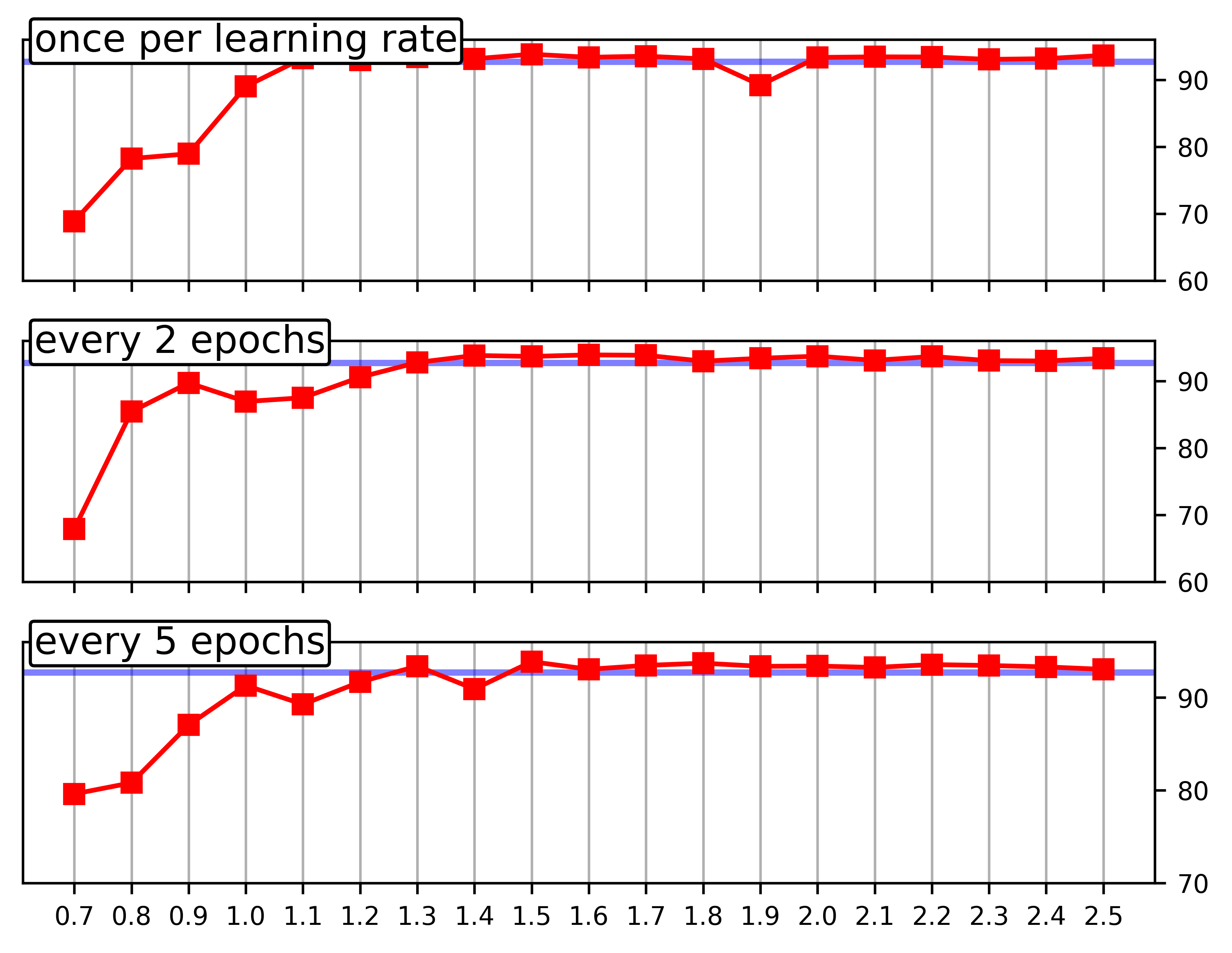}}
  \caption{Performance of the amplified models (red) where layers are selected at different rates compared to mean accuracies of the original models (blue) with no gradient amplification. In each figure, we show the performance (\%) of the models when amplification layers are selected once per learning rate (top), selected every 2 epochs (middle) and selected every 5 epochs (bottom). Horizontal axis refers to the thresholds applied on the normalized gradient rate ($\hat{G_l}$) using case-2 strategy, while vertical axis shows the testing accuracies (\%) of the models. }
  \label{fig:form1_case3_params1_steps}
\end{figure*}

For deeper models, another training strategy $params_2 = [(10, 0.1, 0, 1), \linebreak(100, 0.1, is\_amp, 2), (145, 0.01, is\_amp, 2), (150, 0.01, 0, 1)]$ is employed, in which no amplification is performed for the first 10 epochs and the gradients in the $11^{th}$ epoch are analyzed to identify the layers to perform amplification until $100^{th}$ epoch. At epoch 101, learning rate is reduced to 0.01 and the gradients are analyzed again to determine next set of amplification layers, where the selected layers are amplified for the next 44 epochs and the last 5 epochs are trained without amplification. The number of epochs trained without amplication can be varied at runtime. In our experiments, we just demostrate with either 5 or 20 as our last epochs without amplification. Experiments are performed varying thresholds from 1 to 3 with the step size of 0.25.

We also perform analysis on how frequent the amplification layers need to be varied while training the model. In our training strategy, amplification is applied from 51-100 ($\eta=0.1$) and 101-130 ($\eta=0.01$) epochs. At first, amplification layers are determined on the onset of amplification epochs for each different learning rate and then analysis is done when the amplification layers are changed every 2 epochs from 51-130 and then every 5 epochs.

\section{Results} \label{sec:results}

In our experiments, firstly we analyze simpler models namely, resnet-18, resnet-34 and VGG-19 models using CIFAR-10 dataset and then extend to deeper resnet architectures and also for CIFAR-100 dataset. 

While training these models, a first few epochs are trained normally without any amplification. Then the gradients of the models are analyzed for an epoch by computing normalized gradient rates for all layers.  As mentioned previously, for each layer it determines the rate of fluctuations of weight updates, with 1 corresponding to less fluctuations and 0 for more fluctuations. Since thresholds are measured on normalized gradient rates of layers, values beyond thresholds signify the percentage of layers being amplified and indirectly controls the ratio of amplified layers. Lower the threshold value means larger the ratio of amplified layers and vice versa. In our experiments, thresholds are varied from 0.7 - 2.5 in steps of 0.1 for simpler models and  from 1.0 - 3.0 in steps of 0.25 for deeper models.

\subsection{How frequently should amplification layers be modified/reseleted?}
Analysis is also done on how frequent should these amplification layers be selected by running experiments when amplification layers are changed once per each learning rate, every 2 epochs and 5 epochs. Fig. \ref{fig:form1_case3_params1_steps} shows the performance of vgg-19, resnet-18 and resnet-34 models when these models are amplified using case-2 selection strategy and has similar performance for case-1 selection strategy. It can be observed that the performance improvement does not vary much for a model and one can always fine-tune to determine the best possible layer selection frequency. However, for our further analysis on CIFAR-10 dataset on deeper resnet-50, resnet-101 models, amplification layers are changed once per learning rate for simplicity.

\subsection{Analysis on CIFAR-10 dataset}
\paragraph{Analysis on simpler models:}

\begin{figure*}
  \centering
   {\footnotesize {CIFAR-10 dataset, measure $G$}}\par\medskip
   \vspace*{-0.2cm}
    \subfigure[$VGG-19$\label{fig:Formula_1_VGG-19}]{\includegraphics[width=0.33\textwidth]{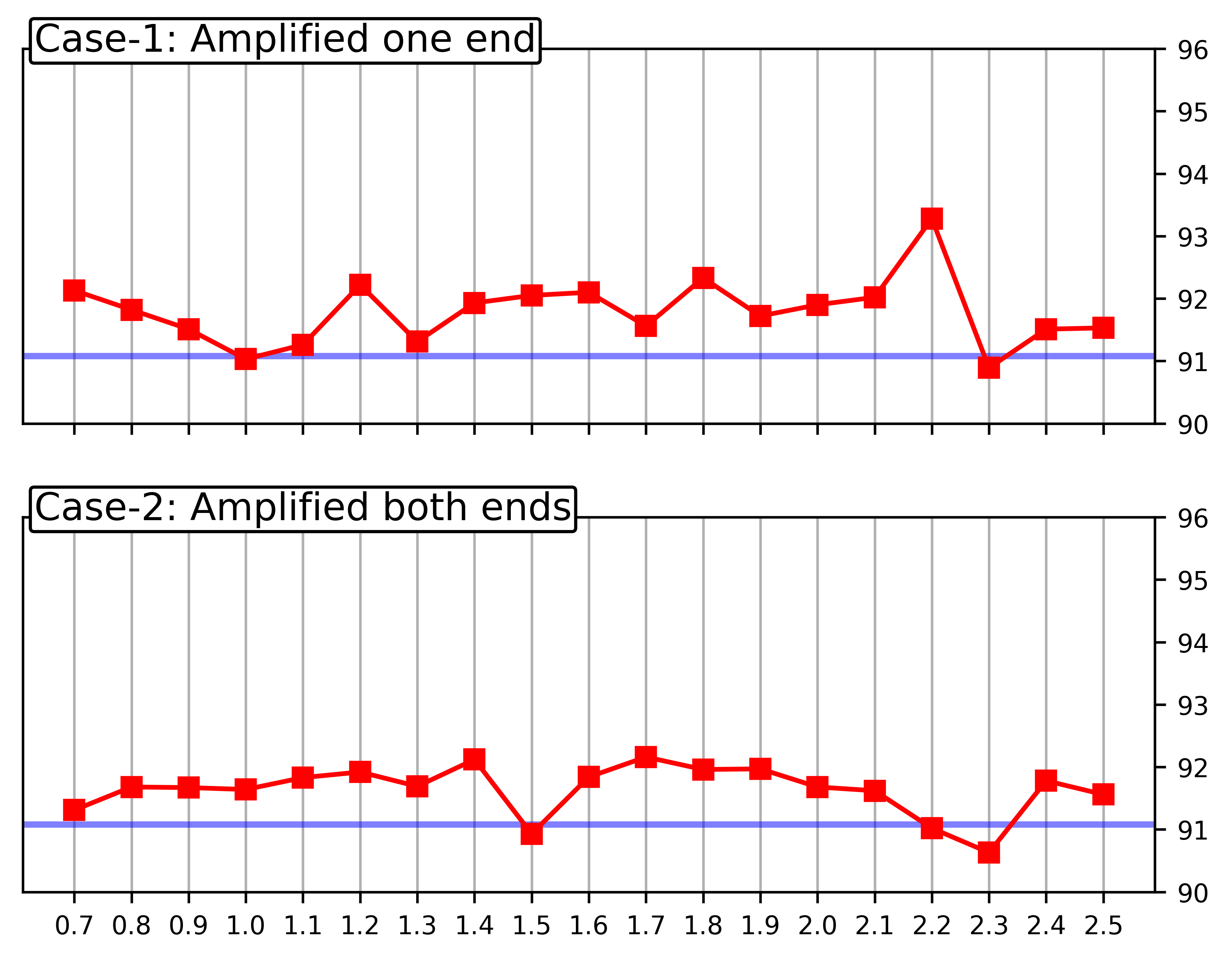}}
    \subfigure[$Resnet-18$\label{fig:Formula_1_resnet_18}]{\includegraphics[width=0.33\textwidth]{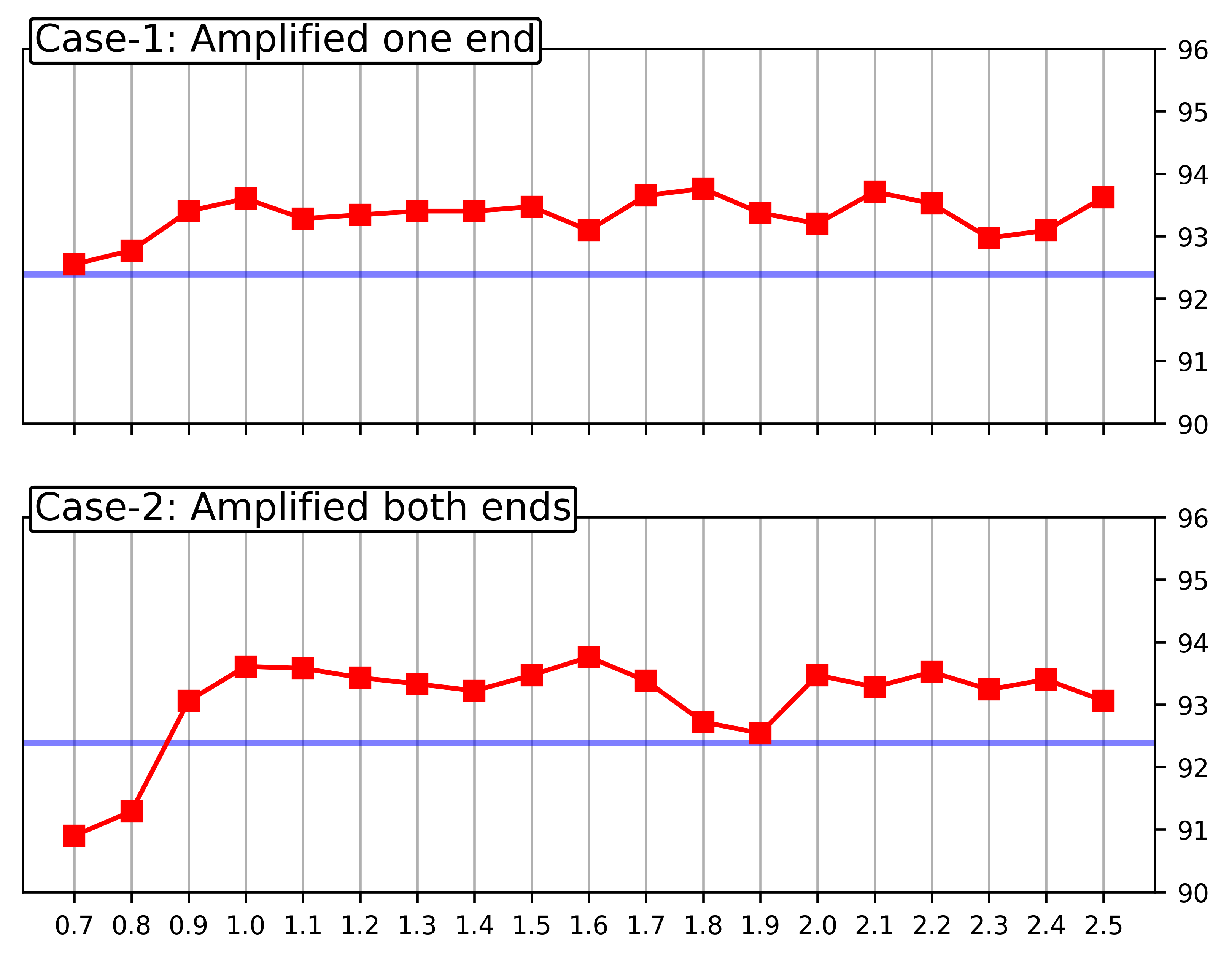}}
    \subfigure[$Resnet-34$\label{fig:Formula_1_resnet_34}]{\includegraphics[width=0.33\textwidth]{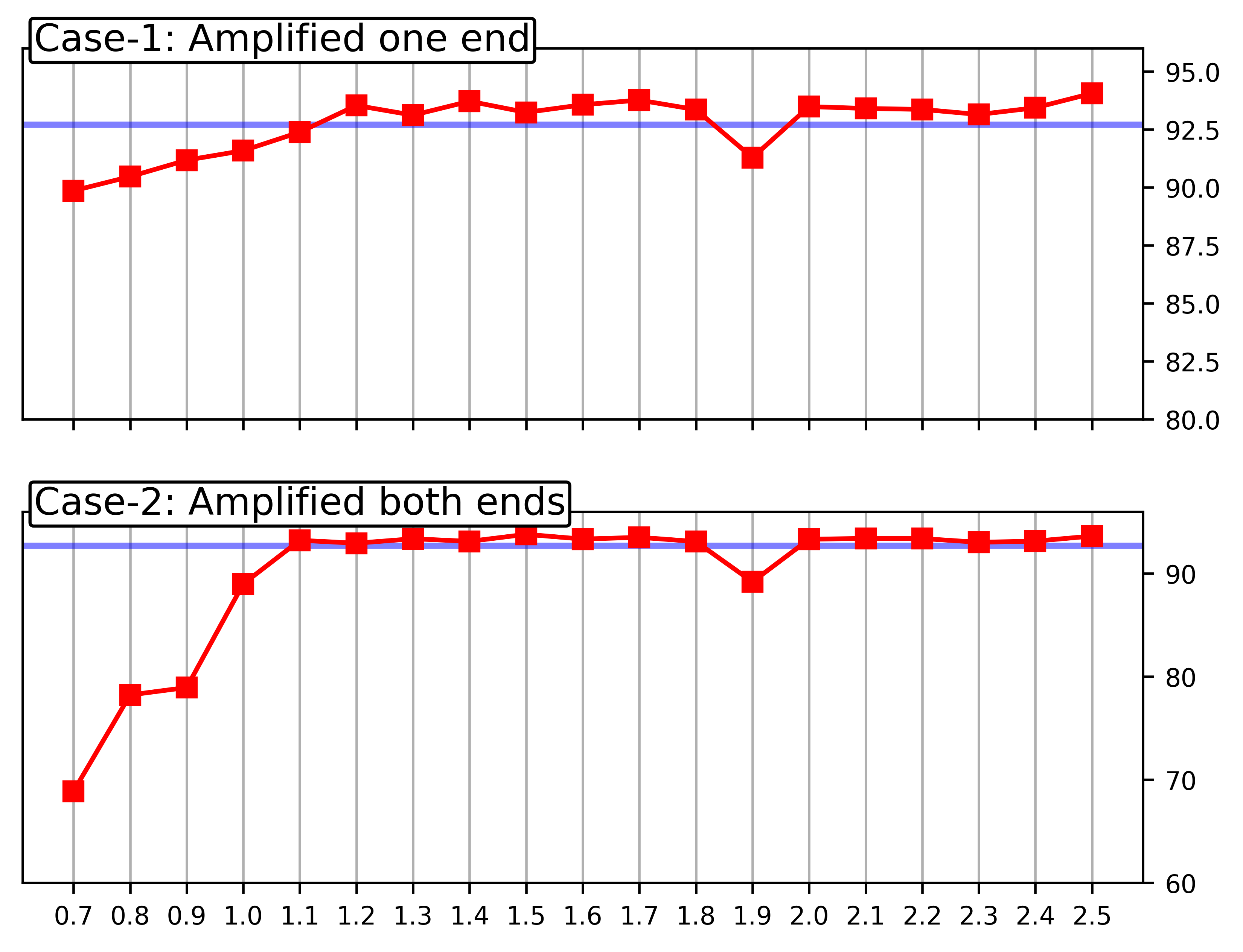}}
  \caption{Performance of the models on CIFAR-10 dataset with amplified models using $G$ applied from epochs 51-130 compared to mean accuracies of the original models(blue) with no gradient amplification. Horizontal axis refers to the thresholds applied on the normalized gradient rate ($\hat{G_l}$) and vertical axis corresponds to testing accuracies (\%) of the models. }
  \label{fig:form1_all_model_cases}
\end{figure*}

\begin{figure*}
  \centering
   {\footnotesize {CIFAR-10 dataset, measure $G'$}}\par\medskip
   \vspace*{-0.2cm}
    \subfigure[VGG-19 \label{fig:Formula_2_VGG-19}]{\includegraphics[width=0.33\textwidth]{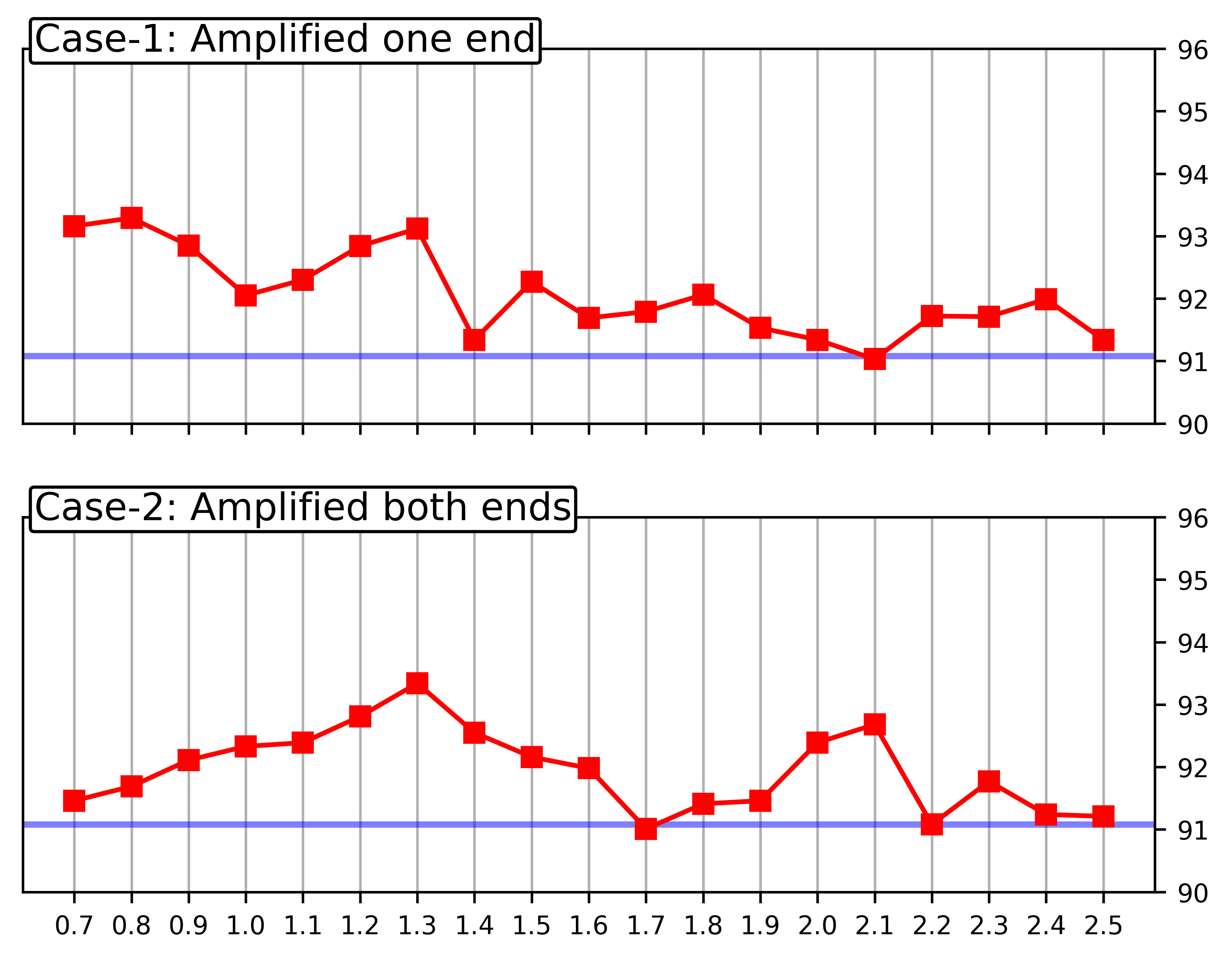}}
    \subfigure[Resnet-18 \label{fig:Formula_2_resnet_18}]{\includegraphics[width=0.33\textwidth]{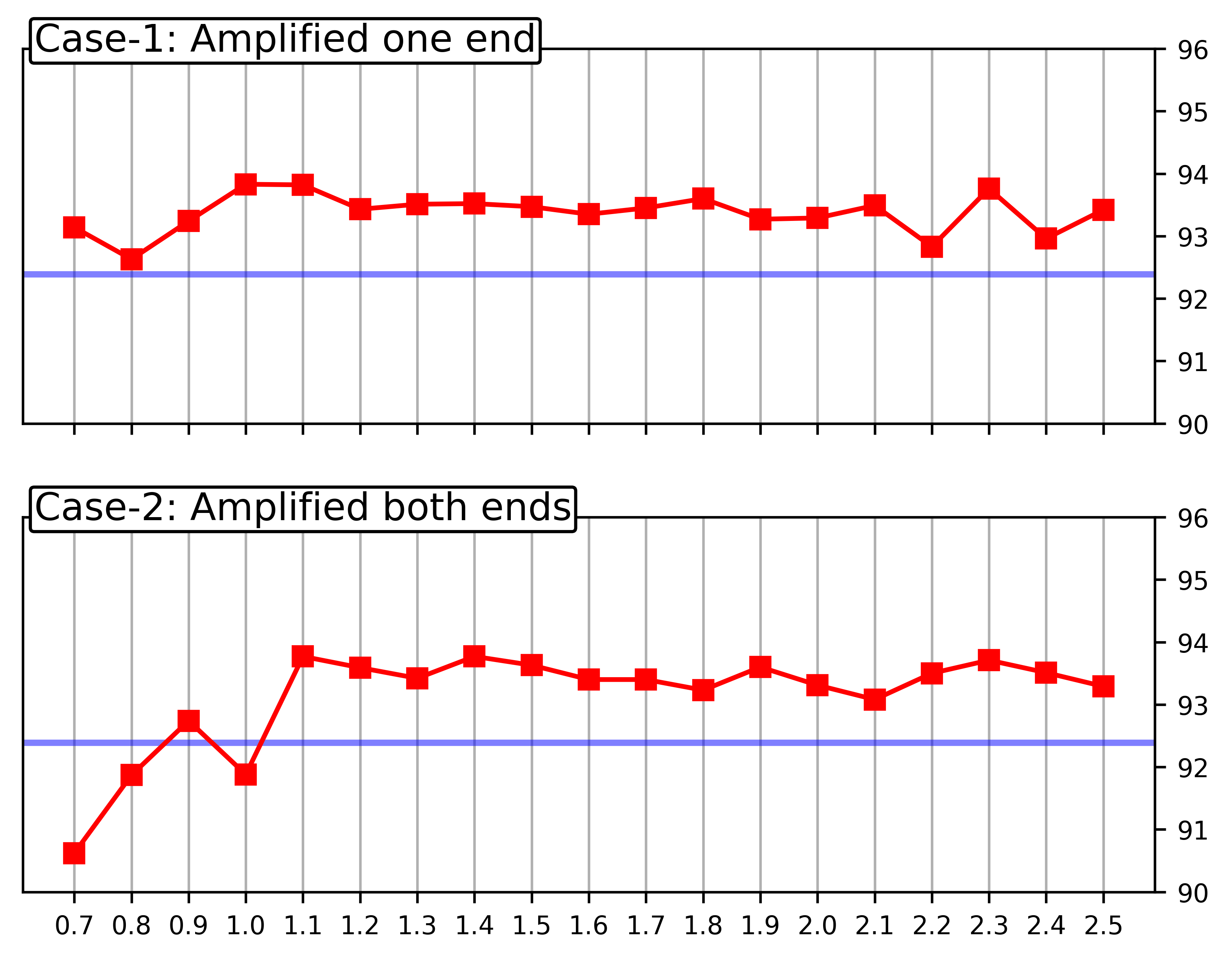}}
    \subfigure[Resnet-34 \label{fig:Formula_2_resnet_34}]{\includegraphics[width=0.33\textwidth]{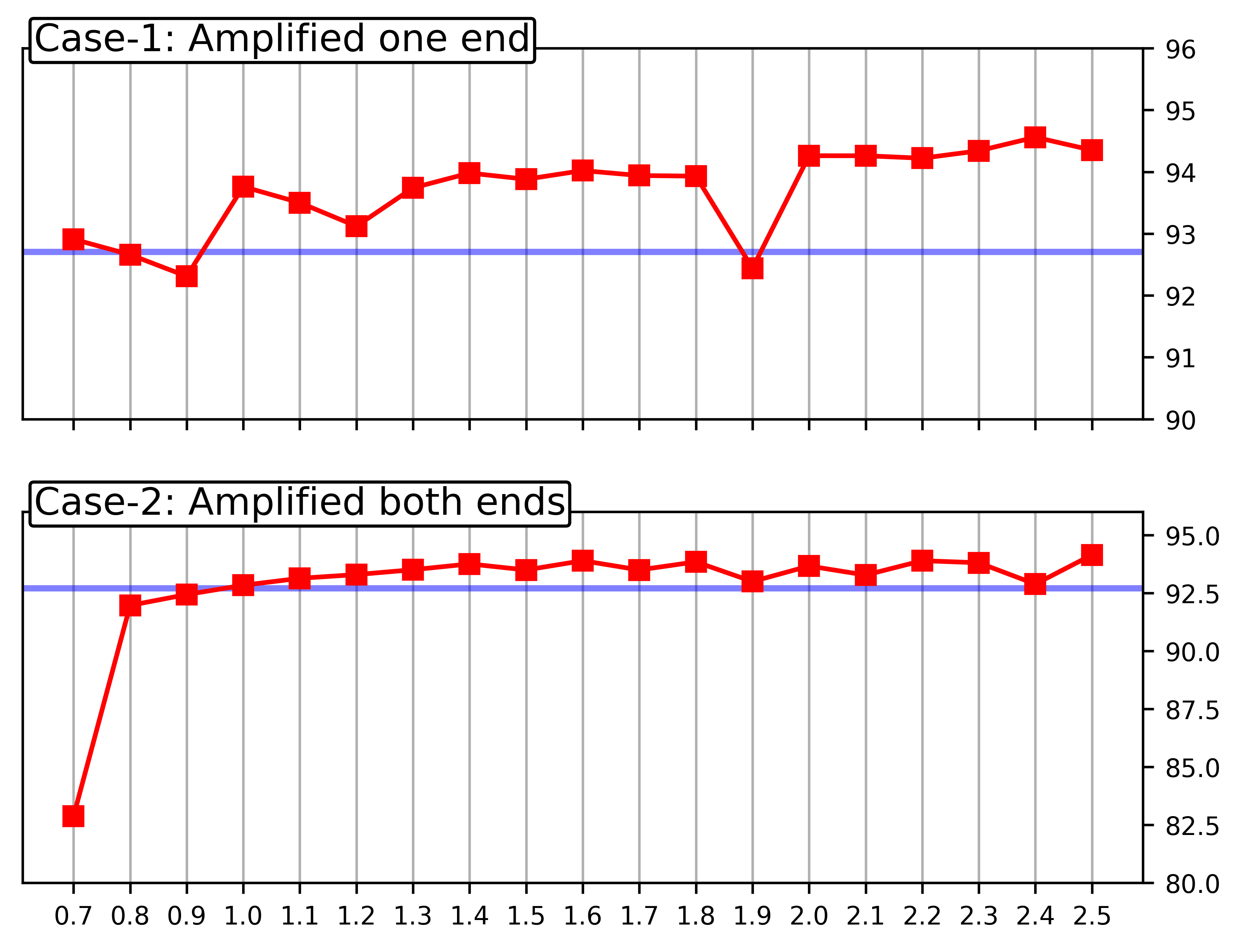}}
  \caption{Testing accuracies \% (Y-axis) of the amplified models (red) using $G'$ compared to mean accuracies of the original models (blue) with no gradient amplification for a range of thresholds (X-axis) applied on the normalized gradient rate ($\hat{G'_l}$) on CIFAR-10 dataset. }
  \label{fig:form2_all_model_cases_cifar10}
\end{figure*}

Fig. \ref{fig:form1_all_model_cases}, \ref{fig:form2_all_model_cases_cifar10} show the performance of VGG-19, resnet-18 and resnet-34 models for a range of thresholds 0.7-2.5 with a step-size of 0.1 when $G$ and $G'$ amplification is applied respectively. For VGG-19 models, when amplified using case-1 and case-2 strategies, accuracies improve for most of the threshold values. Though there is no defined pattern in either case while using $G$, models seem to perform better with lower threshold values less than 1.5 in case of $G'$.  Resnet-18 model perform better while using $G$ or $G'$ formula. In case-1 strategy, models perform for all the threshold values, but for case-2 strategy, models with thresholds greater than 1 perform better than original models. In the case of resnet-34 models, models perform better for thresholds greater than 1.2 and case-1 strategy has better improvements over case-2 in both $G$ and $G'$. Also, the accuracies of the resnet-34 models drop for threshold value of 1.9 in all the cases, otherwise accuracies of the models increase slightly with the increasing thresholds.

\begin{figure*}
  \centering
   {\footnotesize {CIFAR-10 dataset, measure $G$}}\par\medskip
   \vspace*{-0.2cm}
    \subfigure[Resnet-50 \label{fig:Formula_1_resnet_50}]{\includegraphics[width=0.49\textwidth]{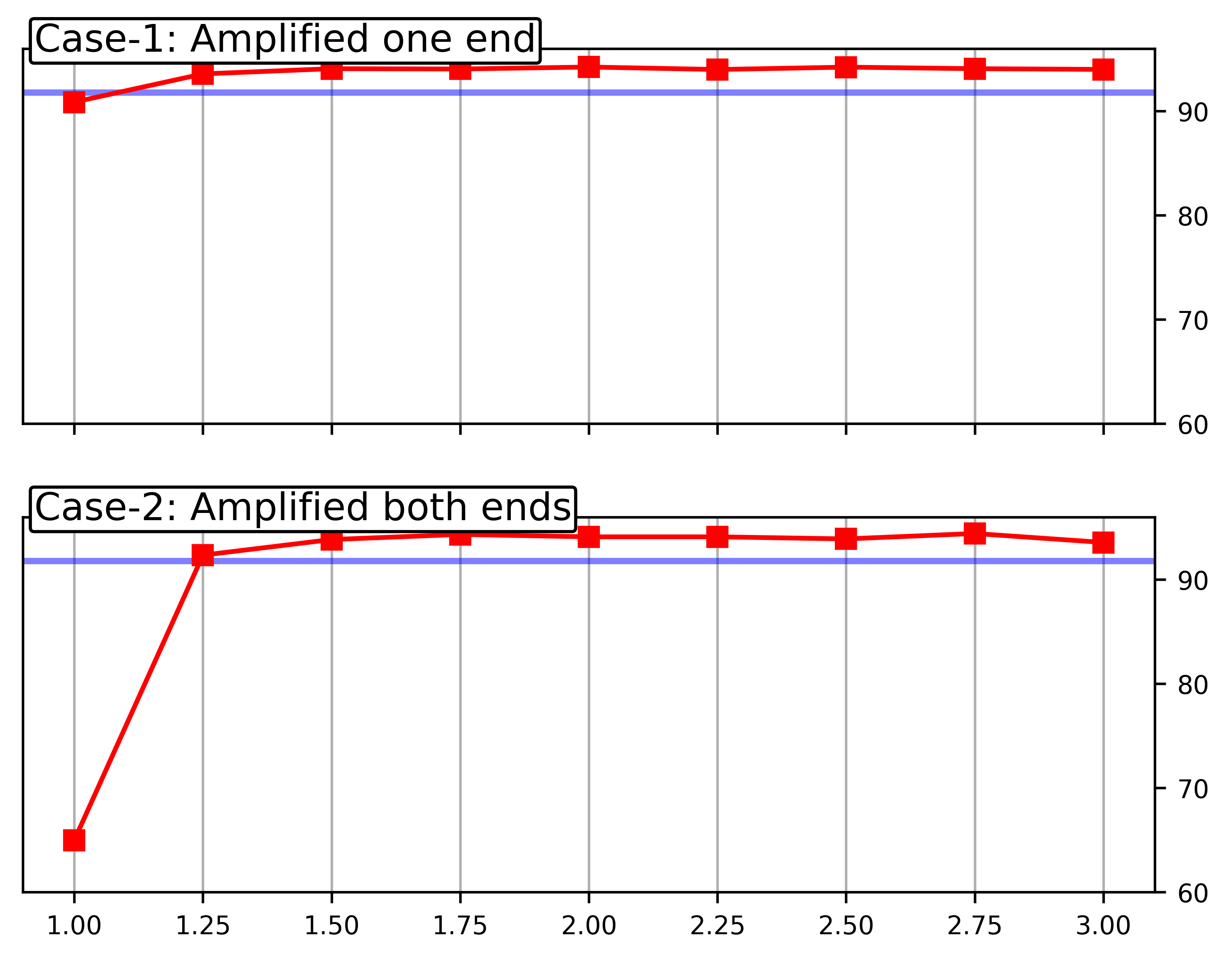}}
    \subfigure[Resnet-101 \label{fig:Formula_1_resnet_101}]{\includegraphics[width=0.49\textwidth]{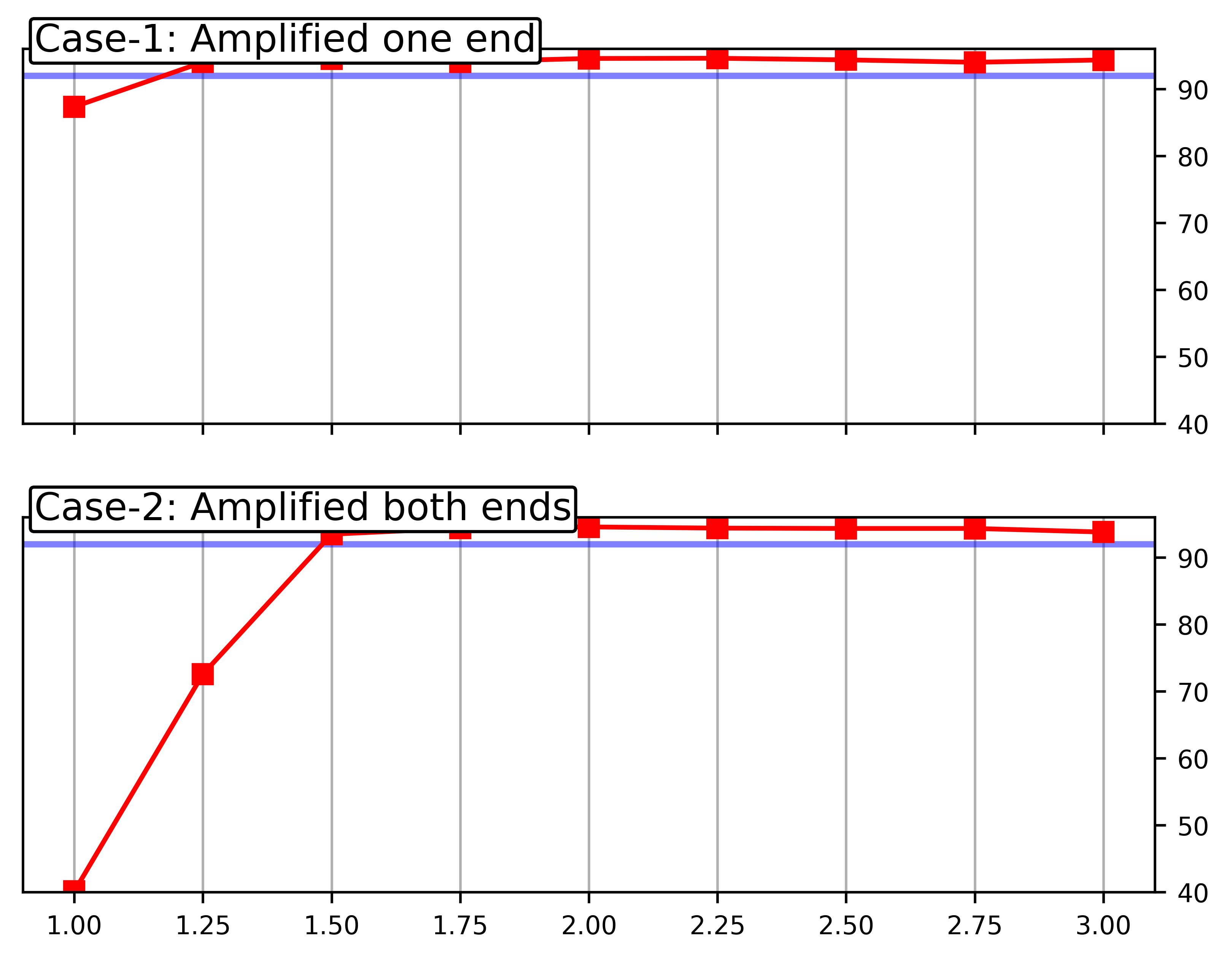}}
  \caption{Testing accuracies  \% (Y-axis) of resnet-50 and resnet-101 models with $G_l$ layer amplification (red) applied from epochs 51-145 compared to mean accuracies of the original models (blue) with no gradient amplification for a range of threshold values (X-axis). }
  \label{fig:form1_deeper_model_cases_cifar10}
\end{figure*}

\begin{figure*}
  \centering
   {\footnotesize {CIFAR-10 dataset, measure $G'$}}\par\medskip
   \vspace*{-0.2cm}
    \subfigure[Resnet-50 \label{fig:Formula_2_resnet_50}]{\includegraphics[width=0.49\textwidth]{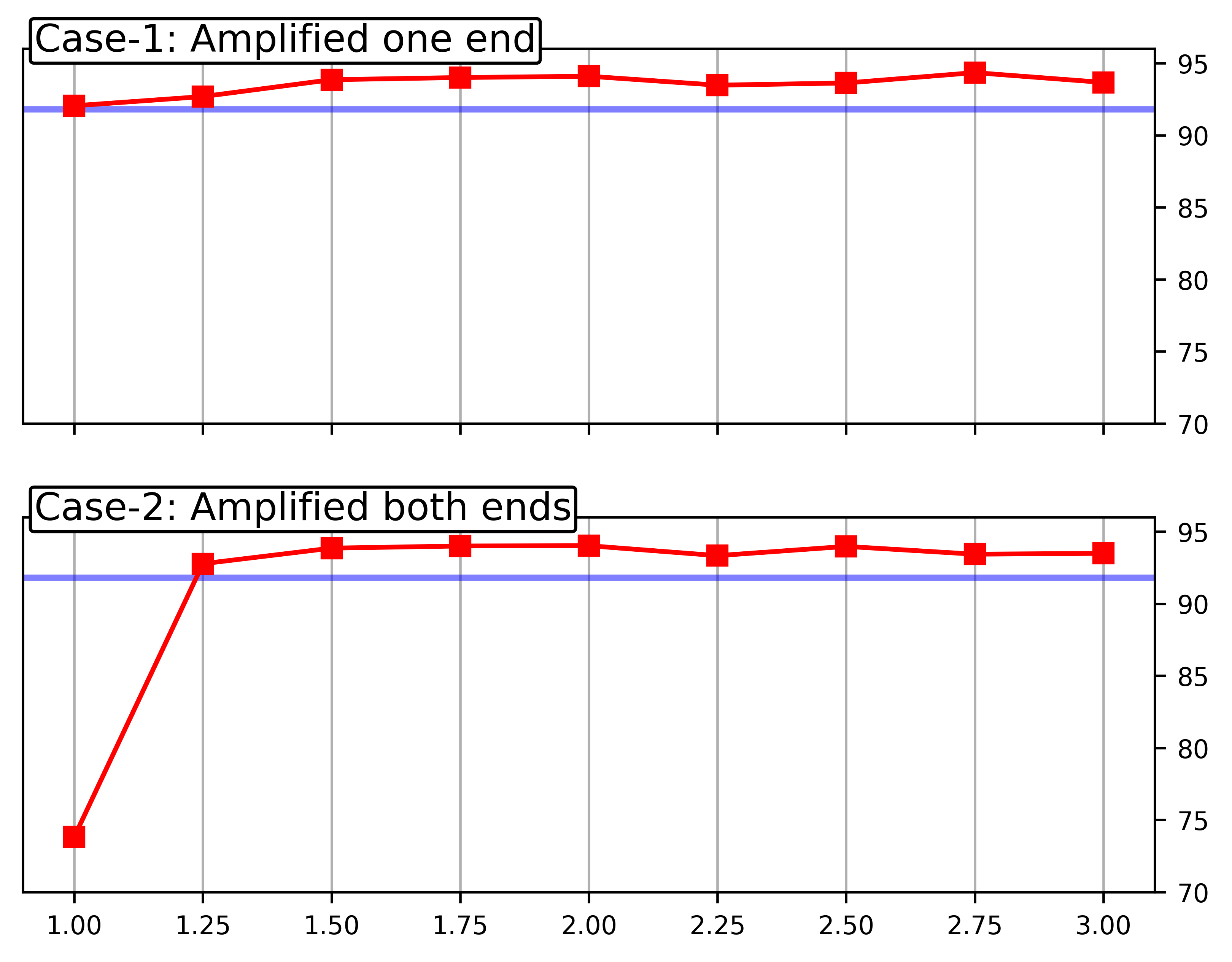}}
    \subfigure[Resnet-101 \label{fig:Formula_2_resnet_101}]{\includegraphics[width=0.49\textwidth]{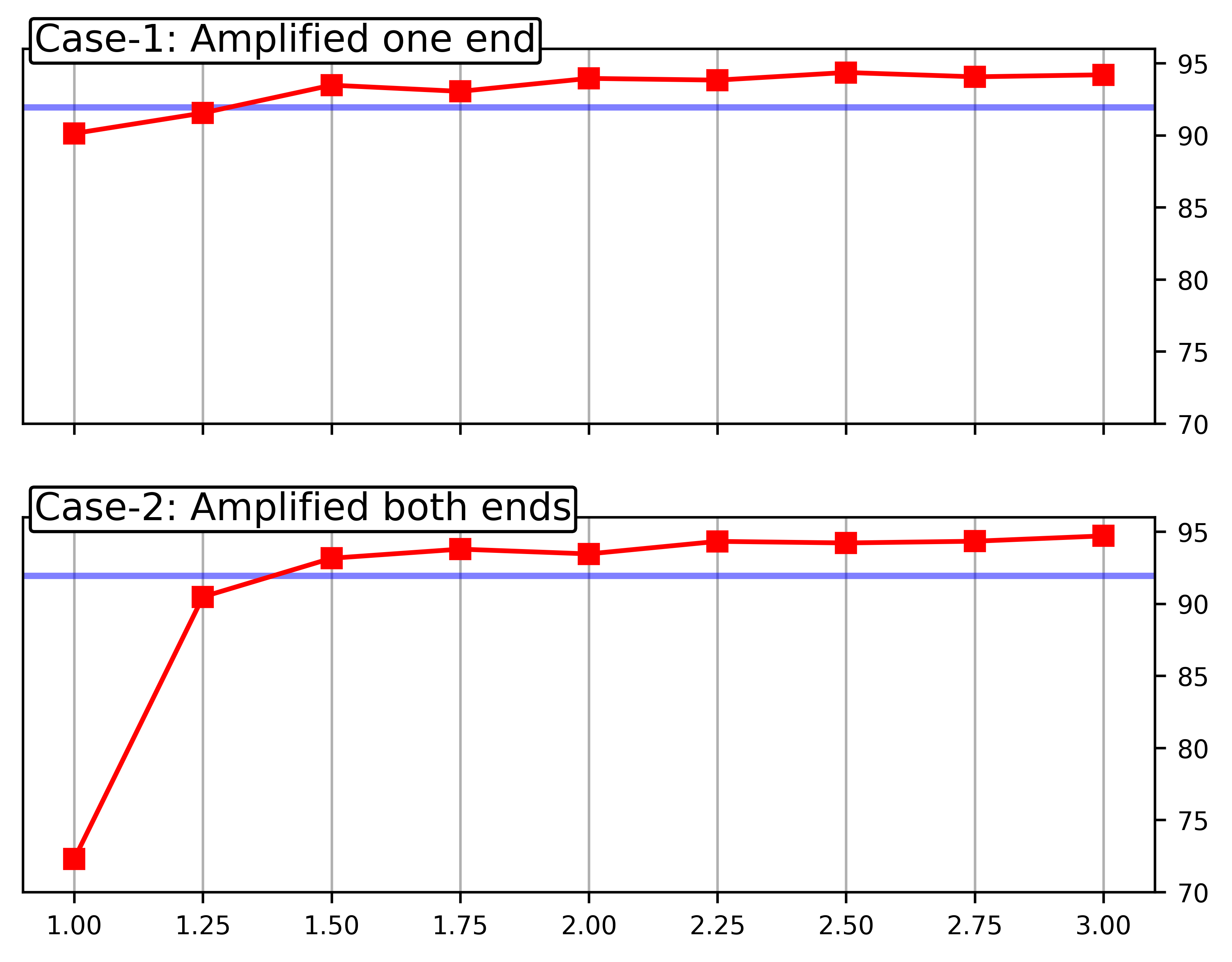}}\quad
  \caption{(CIFAR-10 dataset) Testing accuracies \% (Y-axis) of resnet-50 and resnet-101 models with $G'_l$ layer amplification (red), applied from epochs 51-145, compared to mean accuracies of the original models (blue) with no gradient amplification for a range of threshold values (X-axis). }
  \label{fig:form2_deeper_model_cases_cifar10}
\end{figure*}

 \paragraph{Analysis on deeper models:} As deeper models have longer training times, amplification is performed on reduced thresholds ranging from 1.0 - 3.0 in steps of 0.25. Fig. \ref{fig:form1_deeper_model_cases_cifar10} and \ref{fig:form2_deeper_model_cases_cifar10} show the testing accuracies of the resnet-50 and resnet-101 models for a range of thresholds respectively. For both $G$ and $G'$ while using case-1 strategy, amplified models perform close to original models for lower thresholds and as the threshold value increases amplified models always perform better than original models. Models with case-1 strategy seems to be robust compared to case-2 for lower thresholds. While using case-2 strategy, for lower thresholds, amplified models perform lower than the original models but the performance of the amplified models increases with increasing thresholds. This suggests that deeper models perform better with higher threshold values. Models with case-1 as amplification strategy perform better for thresholds more than $1.25$, and for case-2, models with thresholds $1.5$ perform better.


\subsection{Analysis on CIFAR-100 dataset}
To emphasize the generality of amplification, experiments are also performed on CIFAR-100 dataset. In these experiments, amplification layers are selected once per each learning rate in the training epochs while using previously mentioned training strategies. Experiments are performed with both gradient change measures $G$ and $G'$ using case-1 and case-2 layer selection strategies.

 \paragraph{Analysis on simpler models:}

\begin{figure*}
  \centering
   {\footnotesize {CIFAR-100 dataset, measure $G$}}\par\medskip
   \vspace*{-0.2cm}
    \subfigure[$VGG-19 $\label{fig:Cifar100_Formula_1_VGG_19}]{\includegraphics[width=0.33\textwidth]{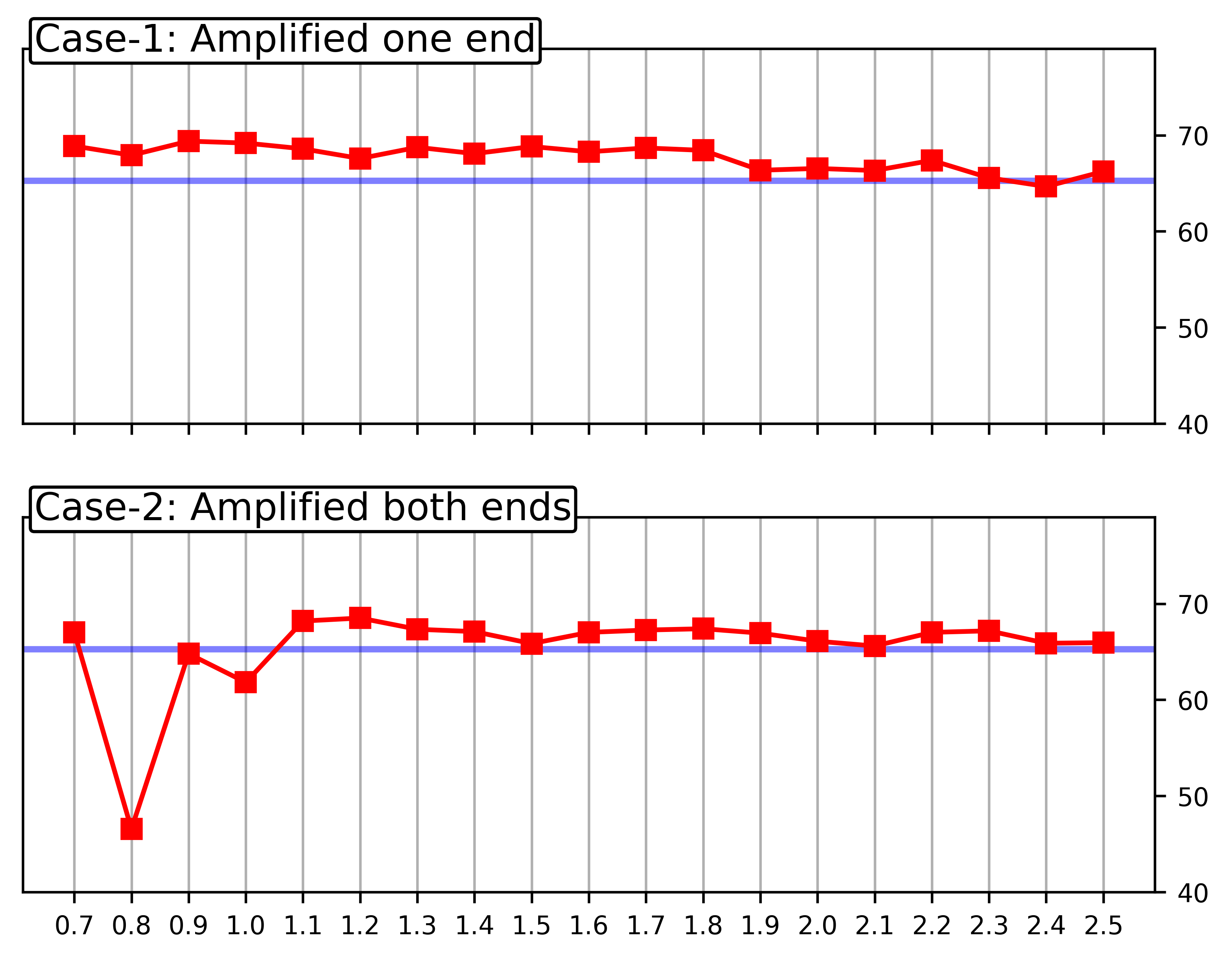}}
    \subfigure[$Resnet-18 $\label{fig:Cifar100_Formula_1_resnet_18}]{\includegraphics[width=0.33\textwidth]{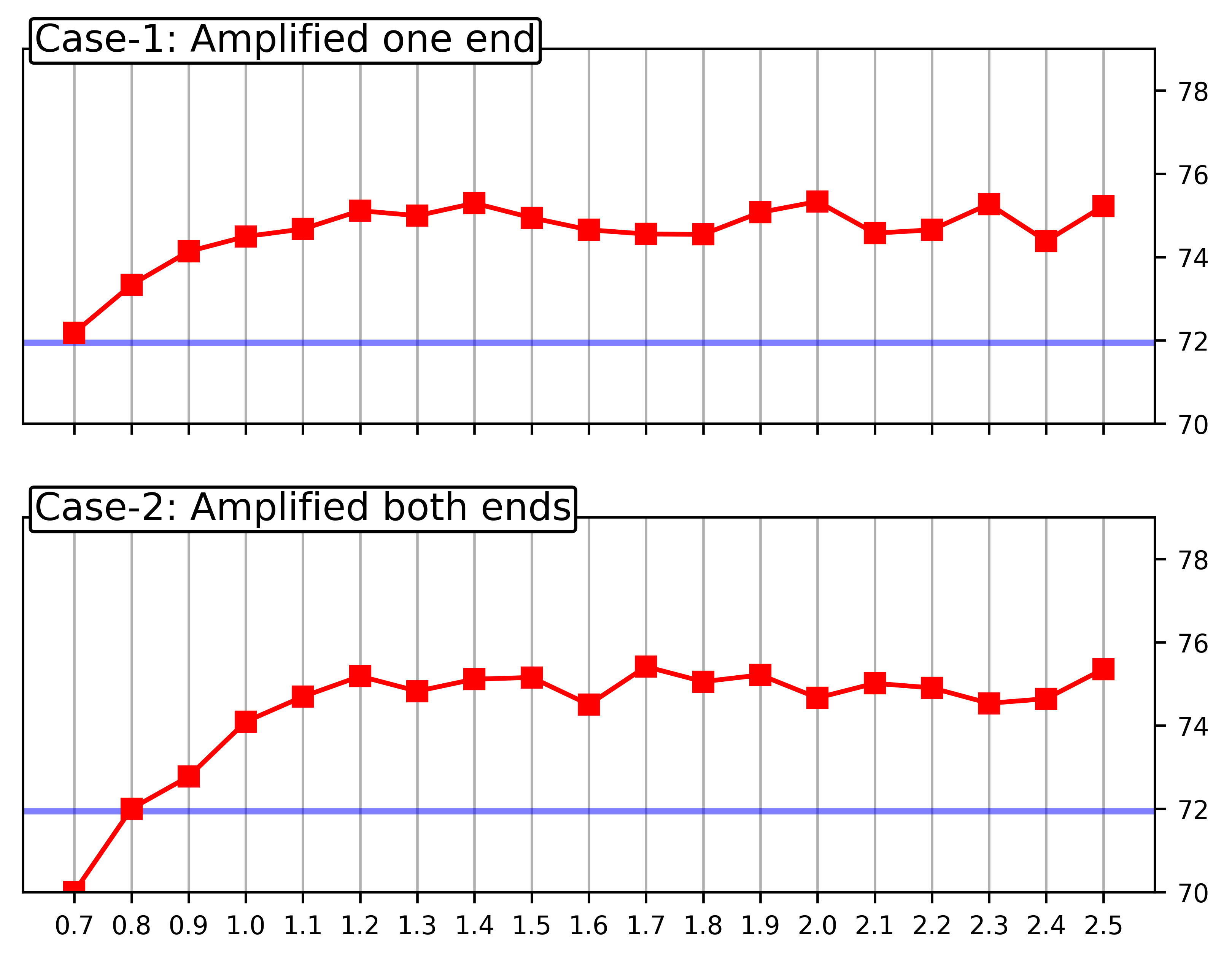}}
    \subfigure[$Resnet-34 $\label{fig:Cifar100_Formula_1_resnet_34}]{\includegraphics[width=0.33\textwidth]{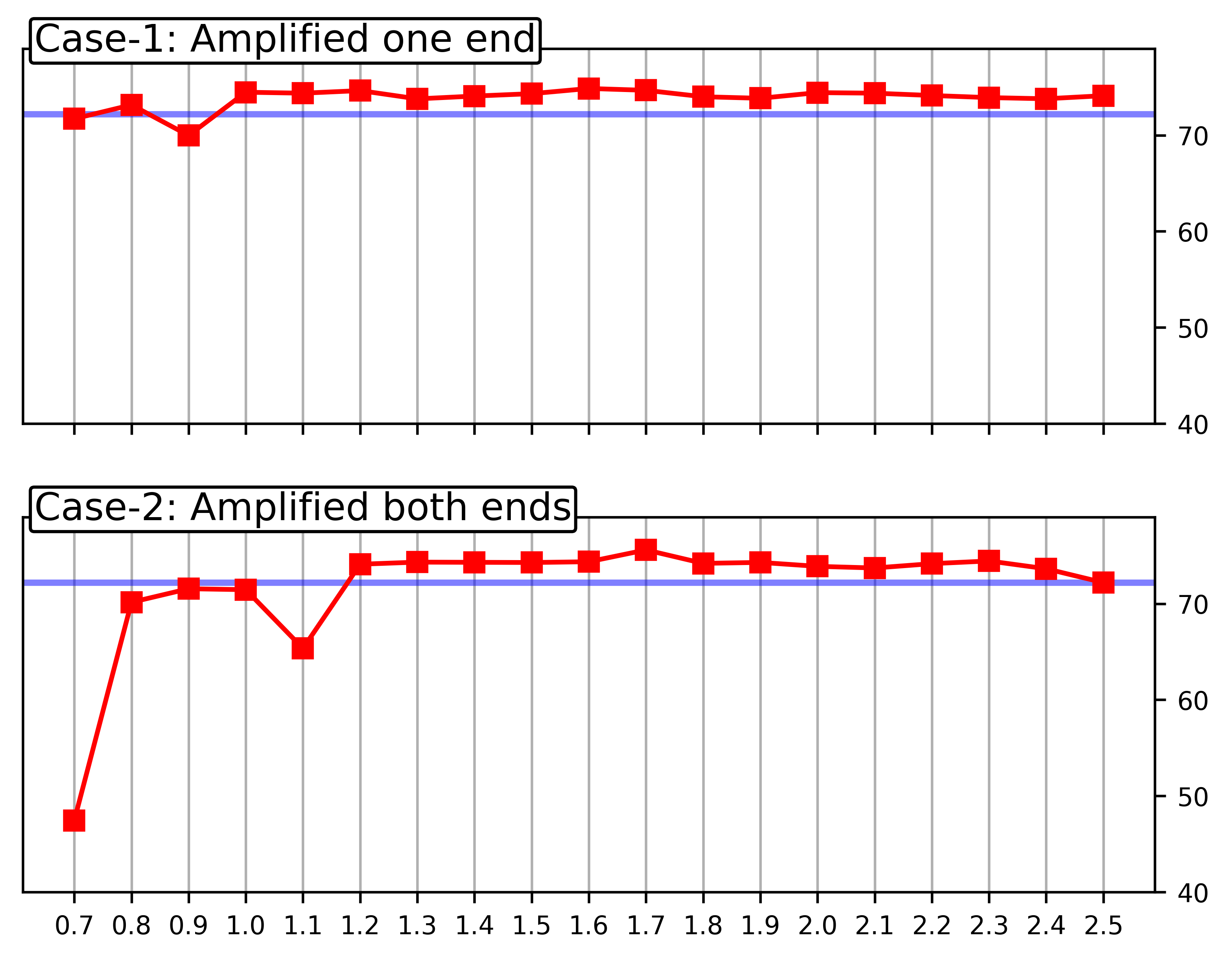}}
  \caption{Performance of the models on CIFAR-100 dataset with amplified models (red) using $G$ applied from epochs 51-130 compared to mean accuracies of the original models (blue) with no gradient amplification. Horizontal axis refers to the thresholds applied on the normalized gradient rate ($\hat{G_l}$) and vertical axis corresponds to testing accuracies (\%) of the models. }
  \label{fig:form1_all_simple_model_cases_cifar100}
\end{figure*}

\begin{figure*}
  \centering
   {\footnotesize {CIFAR-100 dataset, measure $G'$}}\par\medskip
   \vspace*{-0.2cm}
    \subfigure[$VGG-19 $\label{fig:Cifar100_Formula_2_VGG_19}]{\includegraphics[width=0.33\textwidth]{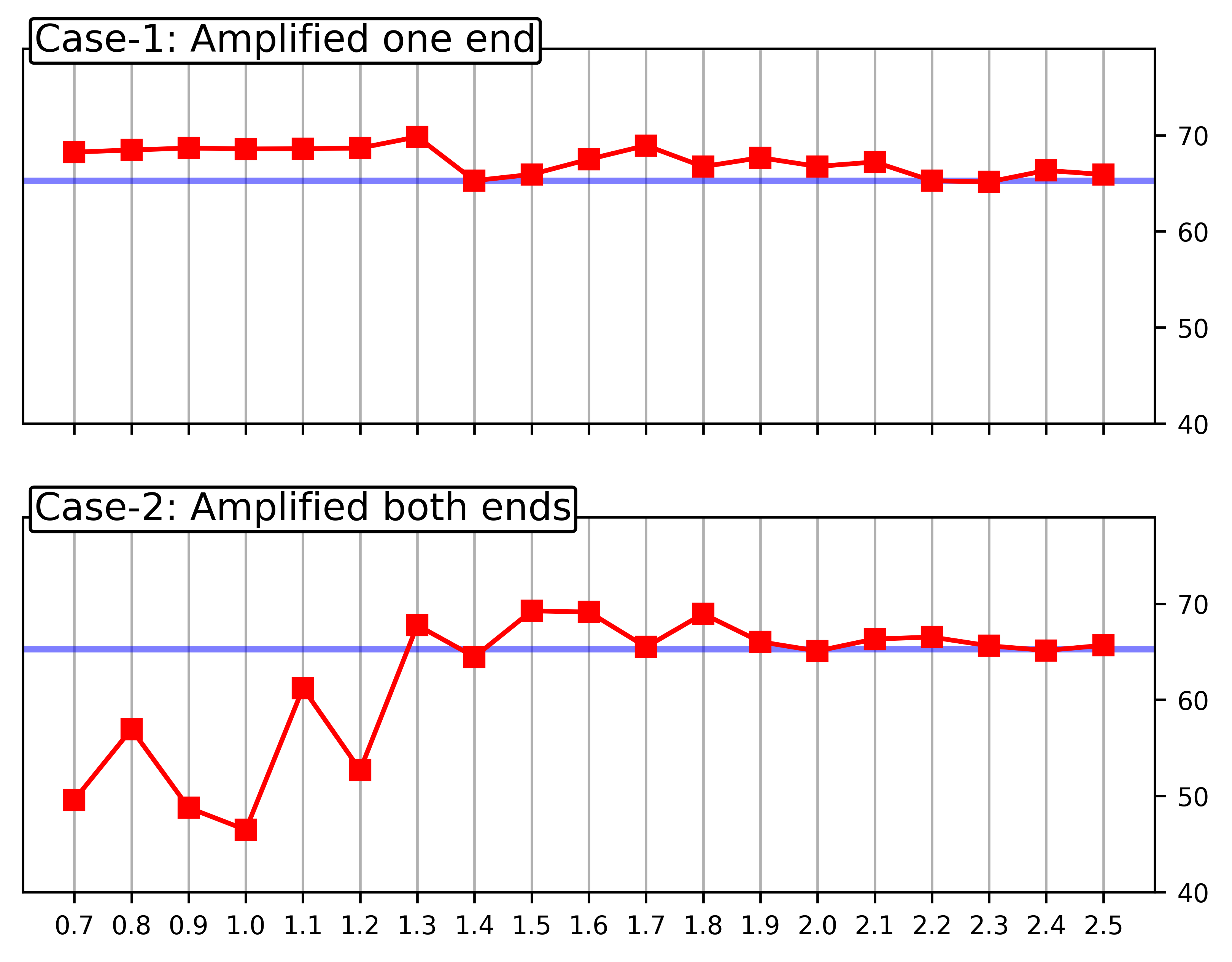}}
    \subfigure[$Resnet-18 $\label{fig:Cifar100_Formula_2_resnet_18}]{\includegraphics[width=0.33\textwidth]{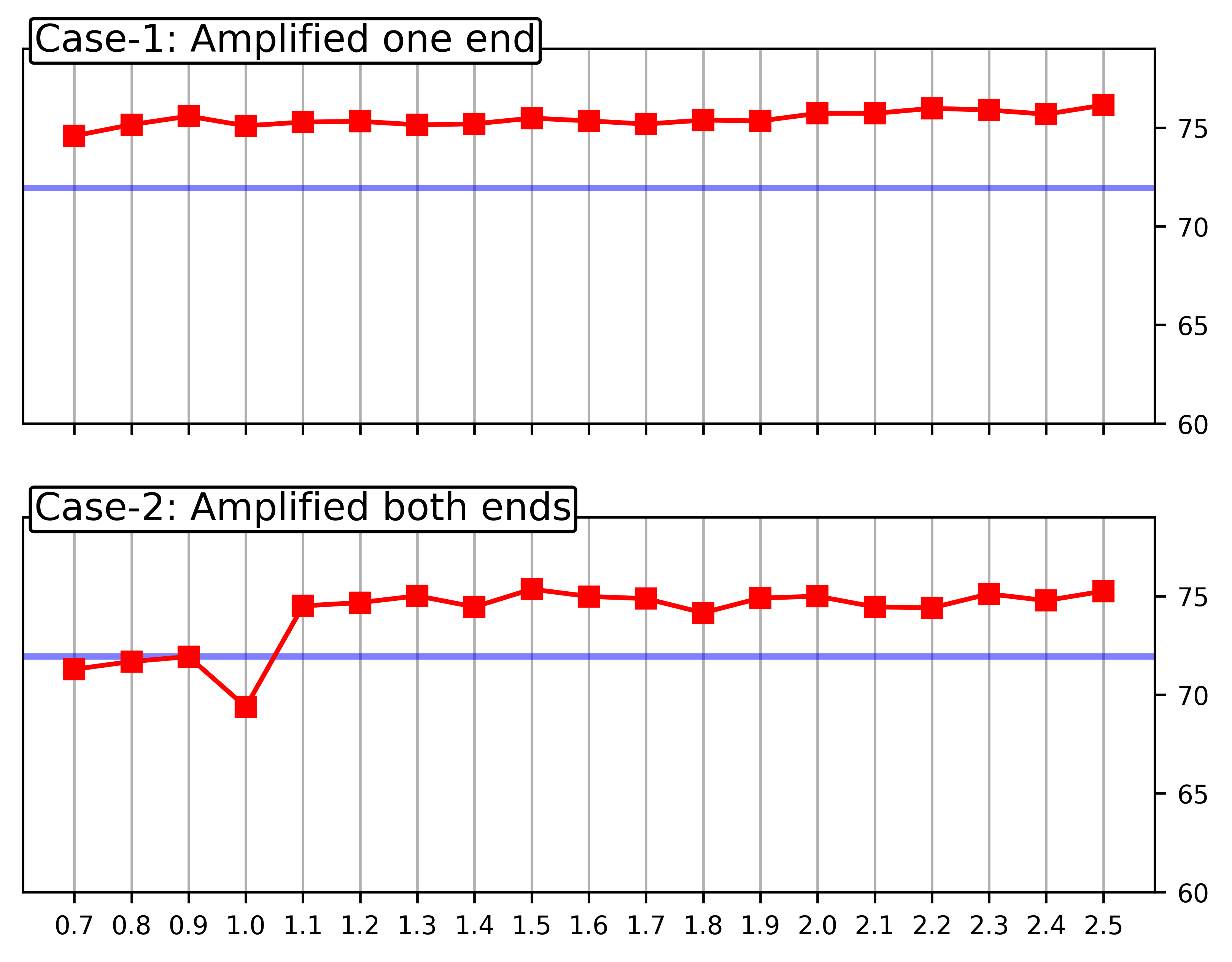}}
    \subfigure[$Resnet-34 $\label{fig:Cifar100_Formula_2_resnet_34}]{\includegraphics[width=0.33\textwidth]{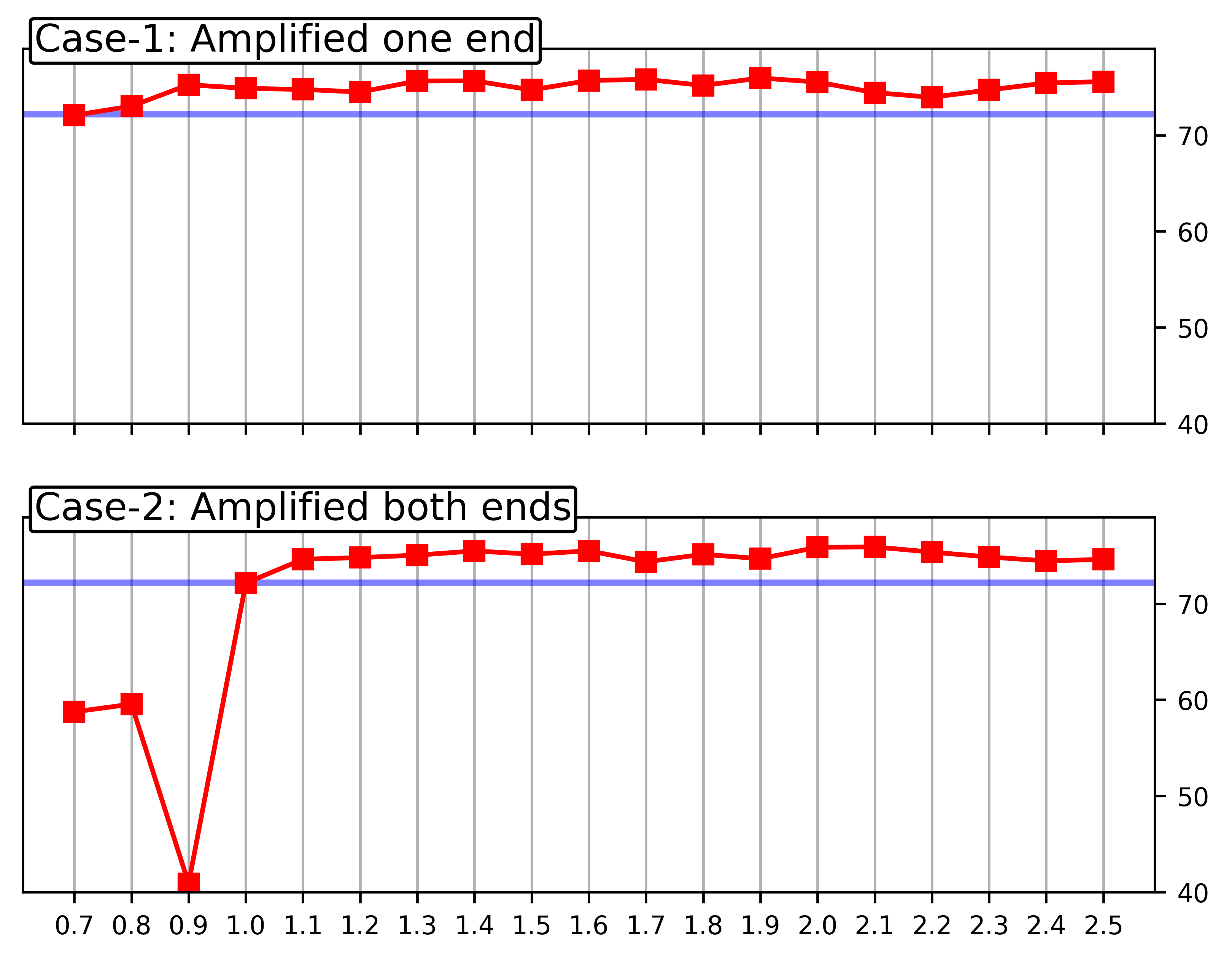}}
  \caption{Testing accuracies \% (Y-axis) of the amplified models (red) using $G'$ compared to mean accuracies of the original models (blue) with no gradient amplification for a range of thresholds (X-axis) applied on the normalized gradient rate ($\hat{G'_l}$) on CIFAR-100 dataset.}
  \label{fig:form2_all_simple_model_cases_cifar100}
\end{figure*}

Fig. \ref{fig:form1_all_simple_model_cases_cifar100}, \ref{fig:form2_all_simple_model_cases_cifar100} show the performance of VGG-19, resnet-18 and resnet-34 models for a range of thresholds 0.7-2.5 with a step-size of 0.1 when $G$ and $G'$ amplification is applied respectively with $params_1=[(50, 0.1, 0, 1), (100, 0.1, is\_amp, 2), (130, 0.01, is\_amp, 2), \linebreak (150, 0.01, 0, 1)]$ as the training strategy. For VGG-19 models, when case-1 amplification method is used, for lower thresholds with both $G$ and $G'$, models perform better at lower threshold values but have similar performance to original models in the case of higher thresholds. When case-2 is used,performance of the models is sensitive to threshold values. Models have lower performance than original models for small thresholds and have similar performance for large thresholds. For intermediate threshold values, it either has better or similar performance to original models. For resnet-18, when case-1 is used, for all the models have better performance than the original models for all the thresholds. Performance of the models increase with the thresholds around 1.5 and then the performance improvement remains the same. While for case-2, models have reduced performance for lower thresholds upto $0.9$ in $G$ and $1.1$ in $G'$ which then increases until $1.5$ and then the improvement remains the same. Resnet-34 models also have similar performance behavior as resnet-18 models maintaining the improved performance with increasing thresholds for case-2. While for case-1, for both $G$ and $G'$, accuracies of the models are lower than the original models (until threshold reaches $1.2$) and are better than original  models after $1.2$. The improvement of the accuracies are maintained with the increasing thresholds.

 \paragraph{Analysis on deeper models:}

\begin{figure*}
  \centering
   {\footnotesize {CIFAR-100 dataset, measure $G'$}}\par\medskip
   \vspace*{-0.2cm}
    \subfigure[Resnet-50 \label{fig:Cifar100_Formula_1_resnet_50}]{\includegraphics[width=0.49\textwidth]{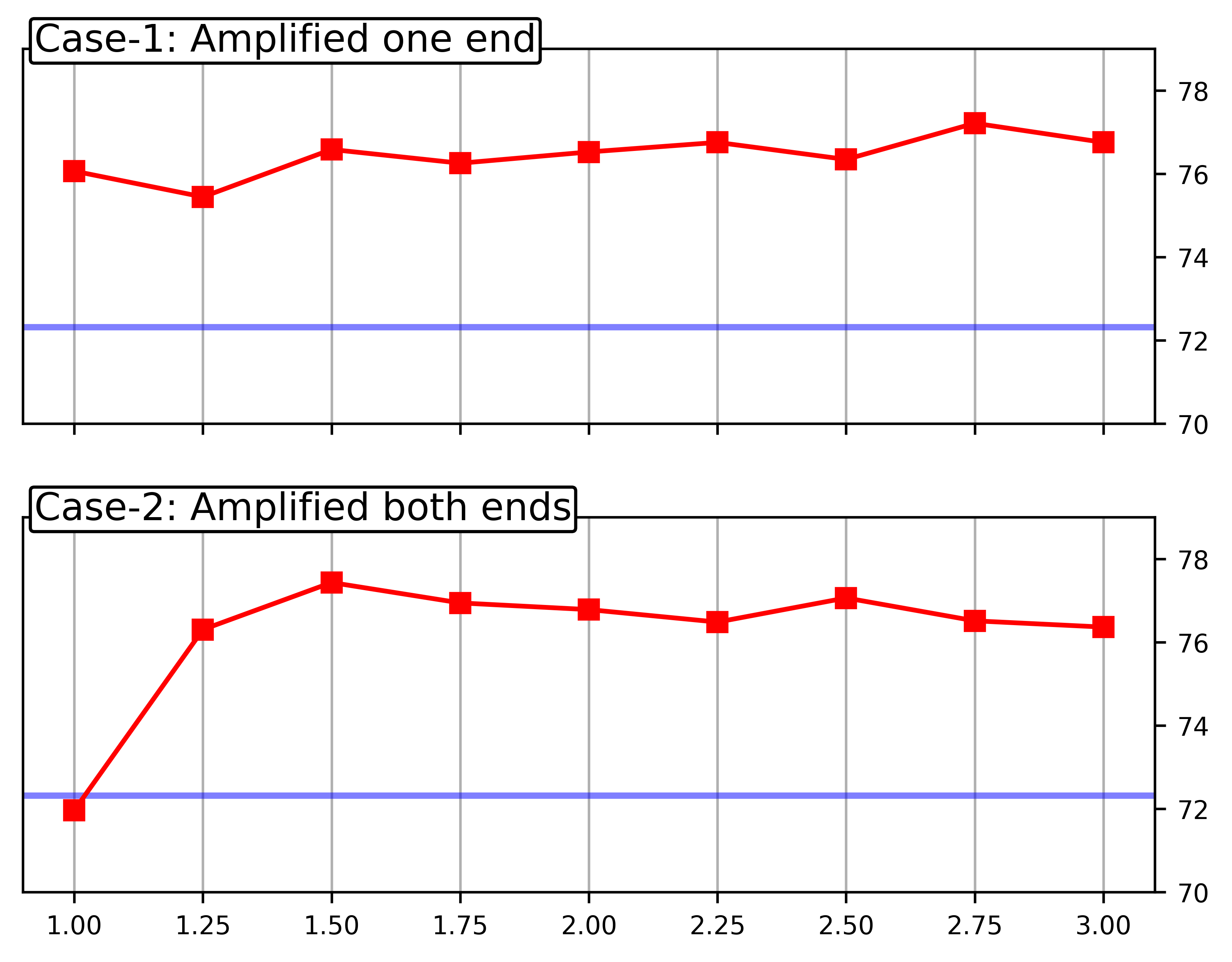}}
    \subfigure[Resnet-101 \label{fig:Cifar100_Formula_1_resnet_101}]{\includegraphics[width=0.49\textwidth]{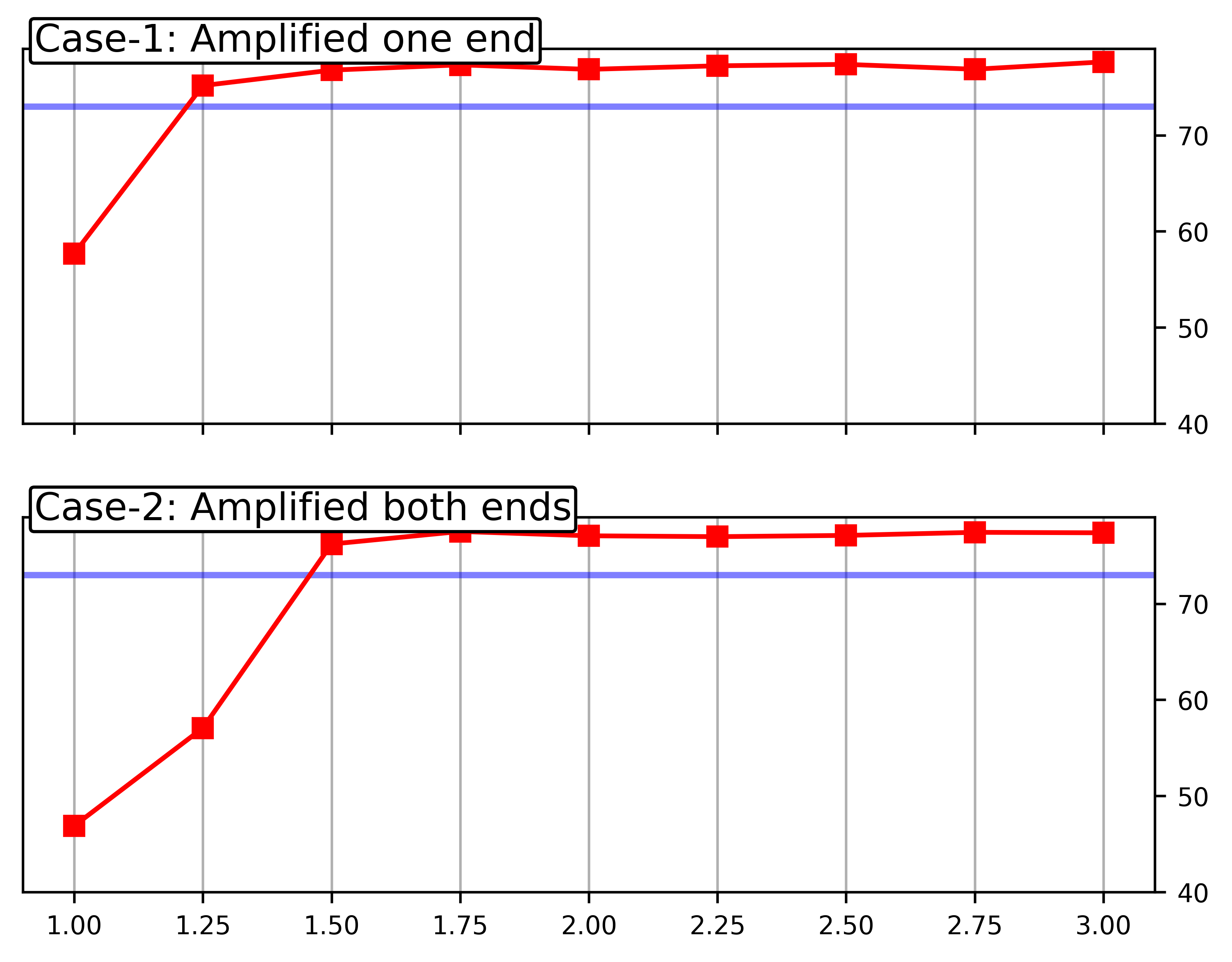}}
  \caption{(CIFAR-100 dataset) Testing accuracies (Y-axis) of resnet-50 and resnet-101 models with $G_l$ layer amplification (red) applied from epochs 51-145 compared to mean accuracies of the original models (blue) with no gradient amplification for a range of threshold values (X-axis).  }
  \label{fig:form1_deeper_model_cases_cifar100}
\end{figure*}

\begin{figure*}
  \centering
   {\footnotesize {CIFAR-100 dataset, measure $G'$}}\par\medskip
   \vspace*{-0.2cm}
    \subfigure[Resnet-50 \label{fig:Cifar100_Formula_2_resnet_50}]{\includegraphics[width=0.49\textwidth]{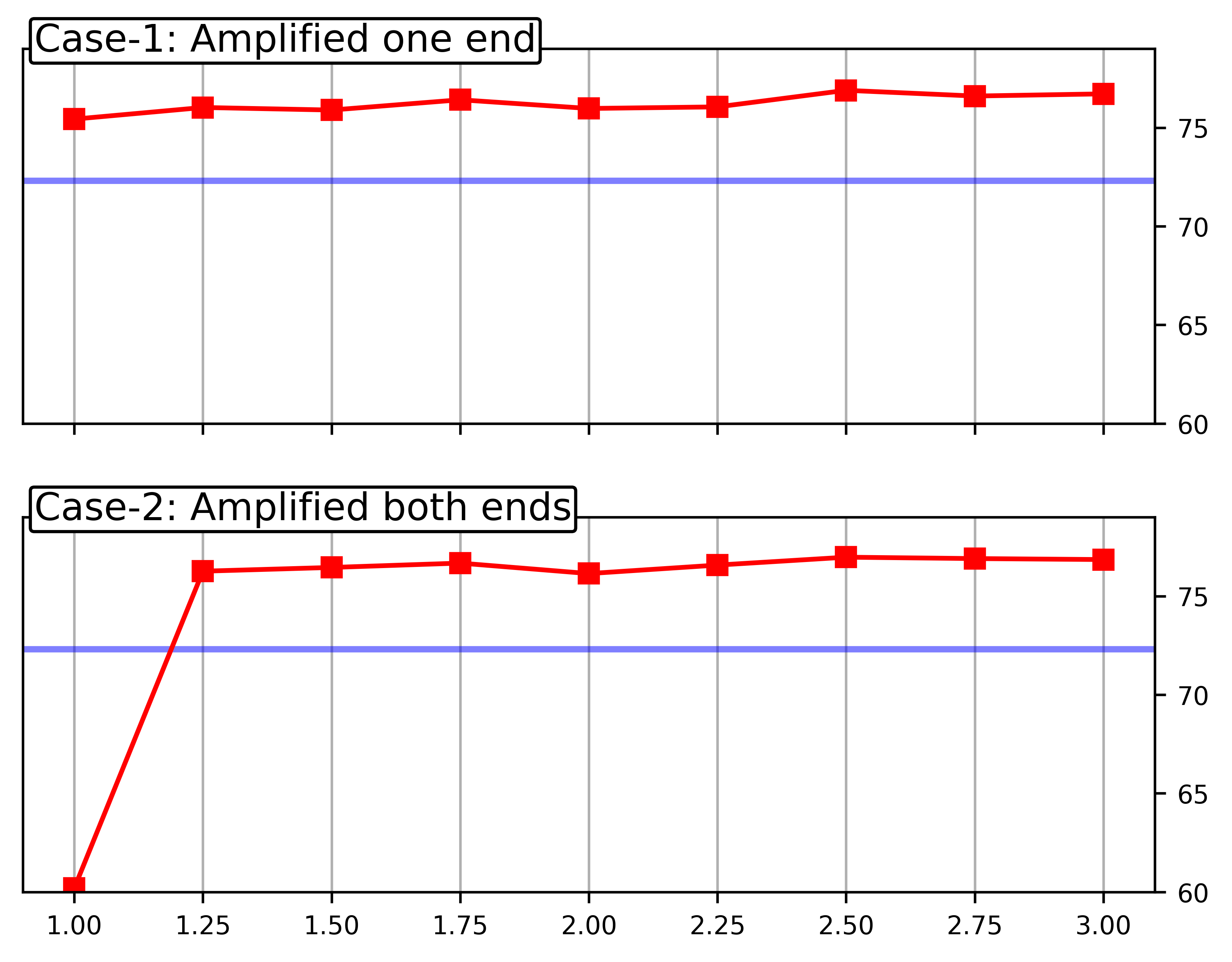}}
    \subfigure[Resnet-101 \label{fig:Cifar100_Formula_2_resnet_101}]{\includegraphics[width=0.49\textwidth]{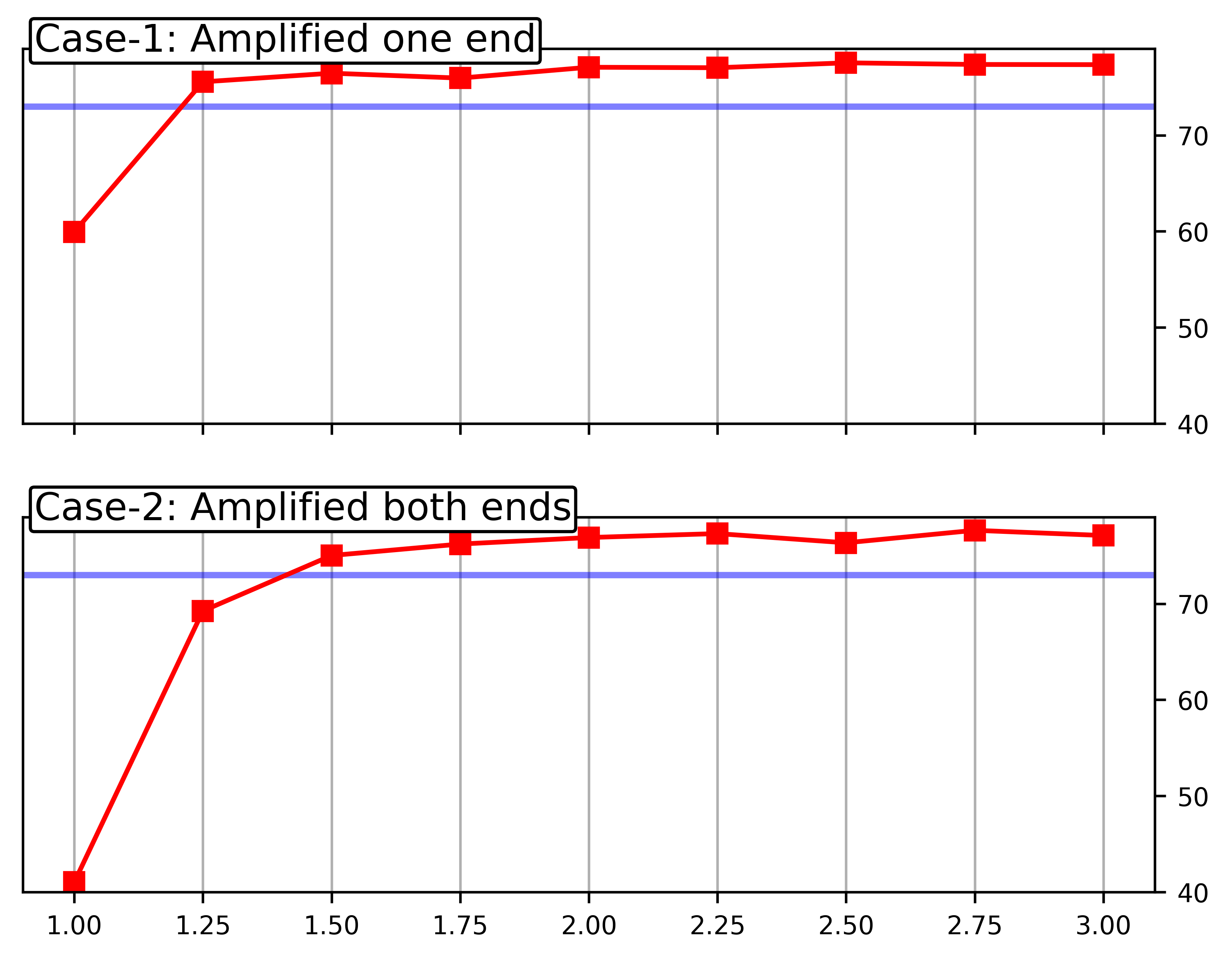}}
  \caption{(CIFAR-100 dataset) Testing accuracies (Y-axis) of resnet-50 and resnet-101 models with $G'_l$ layer amplification (red) applied from epochs 51-145 compared to mean accuracies of the original models (blue) with no gradient amplification for a range of threshold values (X-axis).  }
  \label{fig:form2_deeper_model_cases_cifar100}
\end{figure*}

For deeper networks, resnet-50 and resnet-101, amplification is performed on reduced thresholds ranging from 1.0 - 3.0 in steps of 0.25. Fig. \ref{fig:form1_deeper_model_cases_cifar100} and \ref{fig:form2_deeper_model_cases_cifar100} show the testing accuracies of the resnet-50 and resnet-101 models for a range of thresholds with $params_2=[(10, 0.1, 0, 1), \linebreak (100, 0.1, is\_amp, 2), (145, 0.01, is\_amp, 2),(150, 0.01, 0, 1)]$ as the training strategy. In resnet-50 models, for both $G$ and $G'$ while using case-1 strategy, amplified models always perform better than original models and the improvement of the accuracies almost remain the same across threshold values. While for case-2 strategy, models have lower performance for threshold value $1.00$ and the testing accuracies are better than the original models from threshold $1.25$.  In resnet-101 models,  for both $G$ and $G'$, amplified models perform better from thresholds $1.25$ and $1.50$ respectively while using case-1 and case-2 strategy. Though the improvement in the performance appears the same, testing accuracies improve slowly with increasing thresholds.

\subsection{Comparison of running times}
Table \ref{tab:run_time_cifar10} and \ref{tab:run_time_cifar100} show the mean running times(in minutes) across 10 runs of original models and amplified models for different ratio measures and cases. Training is performed on the GSU high performance cluster with NVIDIA V100 GPUs with only our models running on the GPUs with no other user jobs. Performing amplification while training increases the training times by only 1-3 minutes for all the models in most of the cases.  Therefore, training models performing amplification improves the accuracy of the models maintaining training times closer (less than 2\% increment) to original models.

\begin{table*}[h]
  \caption{ Mean running times(in minutes) of models with $G, G'$ layer-based gradient amplification on CIFAR-10 dataset across 10 iterations. }
  \label{tab:run_time_cifar10}
  \centering
  \resizebox{.99\textwidth}{!}{%

\renewcommand{\arraystretch}{1.4}
  \begin{tabular}{c l l l l l}
    \hline
    &  &\multicolumn{2}{c}{$Ours$ using $\hat{G}$(\textit{min})} &\multicolumn{2}{c}{$Ours$ using $\hat{G'}$(\textit{min})}\\
   \cline{3-6}
   \multirow{-2}{*}{Model} &\multirow{-2}{*}{$Original$ (\textit{min})} & \multicolumn{1}{|c|} {Case-1} & \multicolumn{1}{|c|}{Case-2} & \multicolumn{1}{|c|} {Case-1} & \multicolumn{1}{|c|}{Case-2} \\
    \hline
    $VGG\_19$ & 31.35 \textit{min} $\pm$ 0.45& 31.82 \textit{min} $\pm$ 0.52 (\textbf{1.5\%})& 31.94 \textit{min} $\pm$ 0.45 (\textbf{1.89\%})& 31.87 \textit{min} $\pm$ 0.45 (\textbf{1.66\%})& 31.82 \textit{min} $\pm$ 0.35 (\textbf{1.5\%})\\
    $Resnet\_18$ & 42.15 \textit{min} $\pm$ 0.23& 42.79 \textit{min} $\pm$ 0.45 (\textbf{1.53\%})& 42.73 \textit{min} $\pm$ 0.46 (\textbf{1.38\%})& 42.96 \textit{min} $\pm$ 0.57 (\textbf{1.92\%})& 42.76 \textit{min} $\pm$ 0.44 (\textbf{1.45\%})\\
    $Resnet\_34$ & 66.35 \textit{min} $\pm$ 0.92& 67.73 \textit{min} $\pm$ 0.95 (\textbf{2.09\%})& 67.64 \textit{min} $\pm$ 0.9 (\textbf{1.94\%})& 67.8 \textit{min} $\pm$ 0.82 (\textbf{2.18\%})& 67.68 \textit{min} $\pm$ 0.91 (\textbf{2.01\%})\\
    $Resnet\_50$ & 139.64 \textit{min} $\pm$ 1.97& 139.71 \textit{min} $\pm$ 2.01 (\textbf{0.05\%})& 140.07 \textit{min} $\pm$ 1.59 (\textbf{0.31\%})& 140.63 \textit{min} $\pm$ 1.29 (\textbf{0.71\%})& 141.06 \textit{min} $\pm$ 1.05 (\textbf{1.01\%})\\
    $Resnet\_101$ &224.22 \textit{min} $\pm$ 3.45& 225.48 \textit{min} $\pm$ 2.89 (\textbf{0.56\%})& 226.37 \textit{min} $\pm$ 2.07 (\textbf{0.96\%})& 227.26 \textit{min} $\pm$ 2.12 (\textbf{1.35\%})& 227.78 \textit{min} $\pm$ 2.02 (\textbf{1.58\%})\\

\hline
    \hline
  \end{tabular}
  }
\end {table*}

\begin{table*}[h]
  \caption{ Mean running times(in minutes) of models with $G, G'$ layer-based gradient amplification on CIFAR-100 dataset across 10 iterations. }
  \label{tab:run_time_cifar100}
  \centering
  \resizebox{.99\textwidth}{!}{%

\renewcommand{\arraystretch}{1.4}
  \begin{tabular}{c l l l l l}
    \hline
    &  &\multicolumn{2}{c}{$Ours$ using $\hat{G}$ (\textit{min})} &\multicolumn{2}{c}{$Ours$ using $\hat{G'}$ (\textit{min})}\\
   \cline{3-6}
   \multirow{-2}{*}{Model} &\multirow{-2}{*}{$Original$ (\textit{min}) } & \multicolumn{1}{|c|} {Case-1} & \multicolumn{1}{|c|}{Case-2} & \multicolumn{1}{|c|} {Case-1} & \multicolumn{1}{|c|}{Case-2} \\
    \hline
    $VGG\_19$ & 31.62 \textit{min} $\pm$ 0.48& 32.55 \textit{min} $\pm$ 0.43 (\textbf{2.92\%})& 31.92 \textit{min} $\pm$ 0.49 (\textbf{0.92\%})& 32.16 \textit{min} $\pm$ 0.69 (\textbf{1.7\%})& 31.83 \textit{min} $\pm$ 0.57 (\textbf{0.64\%})\\
    $Resnet\_18$ & 42.08 \textit{min} $\pm$ 0.47& 42.83 \textit{min} $\pm$ 0.64 (\textbf{1.78\%})& 42.24 \textit{min} $\pm$ 0.69 (\textbf{0.39\%})& 43.44 \textit{min} $\pm$ 2.65 (\textbf{3.23\%})& 42.19 \textit{min} $\pm$ 0.57 (\textbf{0.26\%})\\
    $Resnet\_34$ & 66.64 \textit{min} $\pm$ 0.84& 67 \textit{min} $\pm$ 0.89 (\textbf{0.54\%})& 66.99 \textit{min} $\pm$ 0.96 (\textbf{0.53\%})& 69.84 \textit{min} $\pm$ 6.11 (\textbf{4.8\%})& 66.88 \textit{min} $\pm$ 0.9 (\textbf{0.36\%})\\
    $Resnet\_50$ & 139.32 \textit{min} $\pm$ 1.18& 140.64 \textit{min} $\pm$ 0.51 (\textbf{0.95\%})& 140.21 \textit{min} $\pm$ 1.7 (\textbf{0.64\%})& 140.68 \textit{min} $\pm$ 0.51 (\textbf{0.97\%})& 140.79 \textit{min} $\pm$ 0.68 (\textbf{1.06\%})\\
    $Resnet\_101$ & 223.88 \textit{min} $\pm$ 2.69& 226.35 \textit{min} $\pm$ 3.05 (\textbf{1.10\%})& 226.28 \textit{min} $\pm$ 1.87 (\textbf{1.07\%})& 225.09 \textit{min} $\pm$ 3.23 (\textbf{0.54\%})& 226.12 \textit{min} $\pm$ 2.18 (\textbf{1.00\%})\\

\hline
    \hline
  \end{tabular}
  }
\end {table*}

\begin{figure*}
  \centering 
   {\footnotesize {Best models on CIFAR-10 dataset with measure $G$}}\par\medskip
   \vspace*{-0.2cm}
    \subfigure[VGG-19 \label{fig:cifar10_formula_1_best_epoch-vgg-19}]{\includegraphics[width=0.2\textwidth]{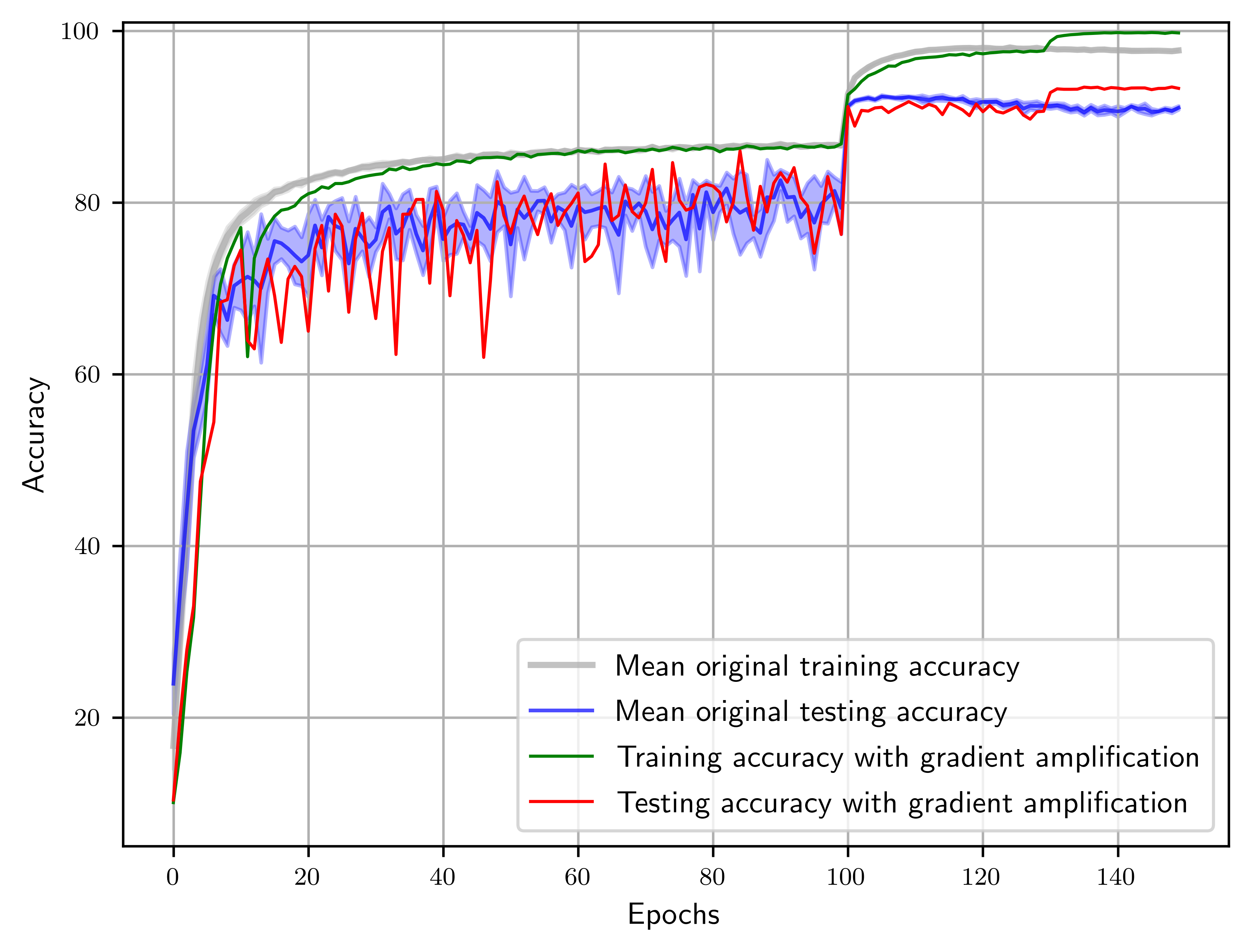}}
    \subfigure[Resnet-18 \label{fig:cifar10_formula_1_best_epoch-resnet-18}]{\includegraphics[width=0.2\textwidth]{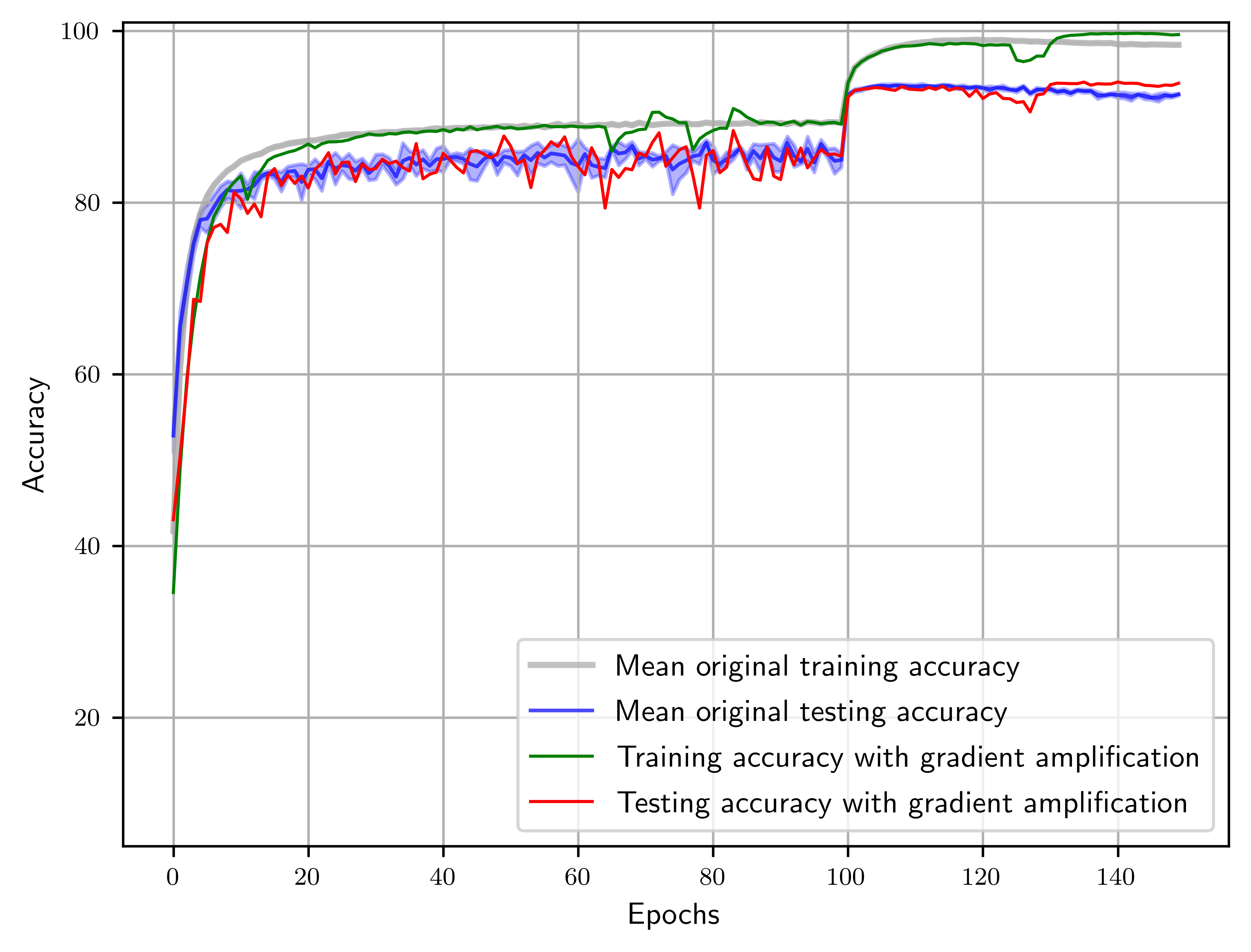}}
    \subfigure[Resnet-34 \label{fig:cifar10_formula_1_best_epoch-resnet-34}]{\includegraphics[width=0.2\textwidth]{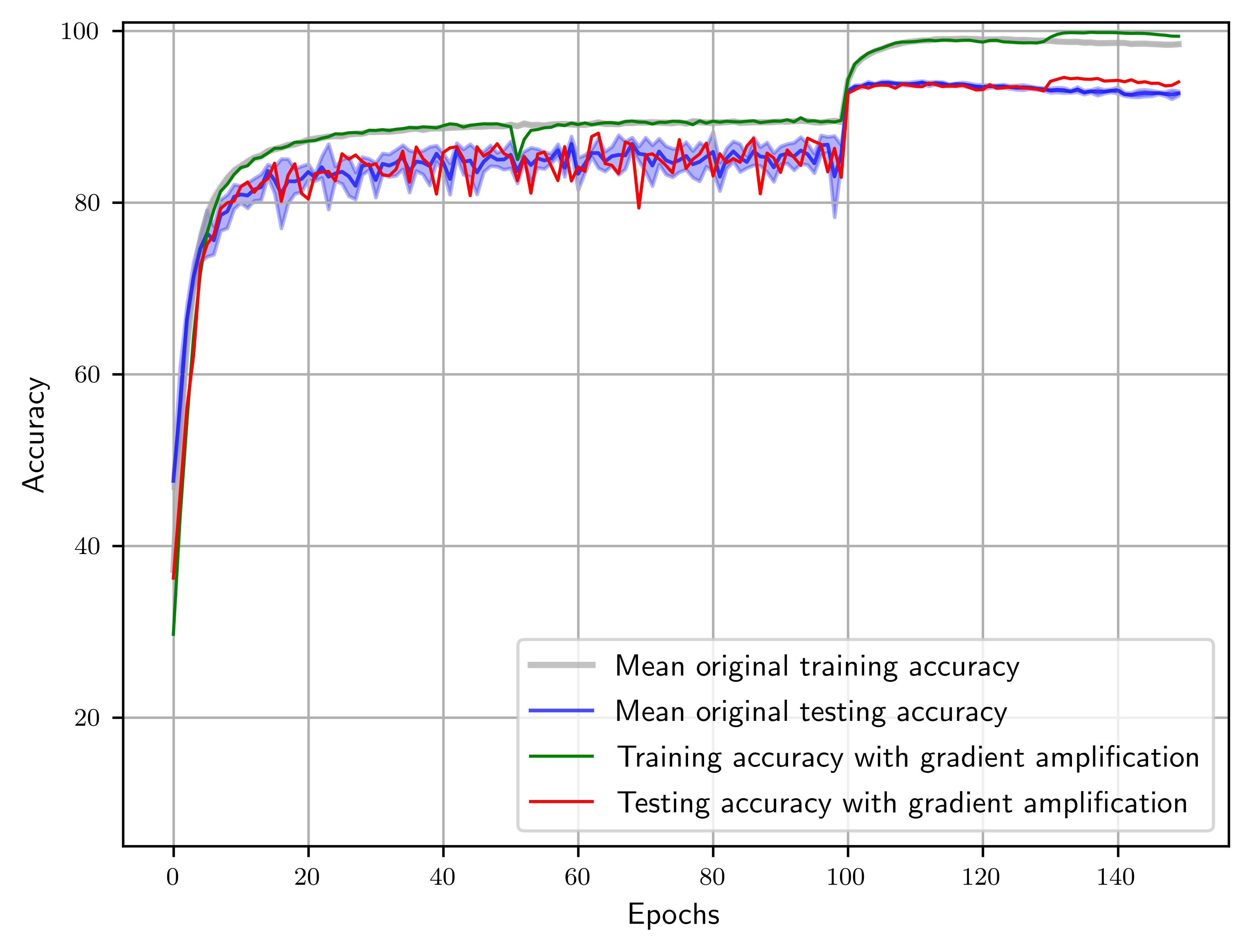}}
    \subfigure[Resnet-50 \label{fig:cifar10_formula_1_best_epoch-resnet-50}]{\includegraphics[width=0.2\textwidth]{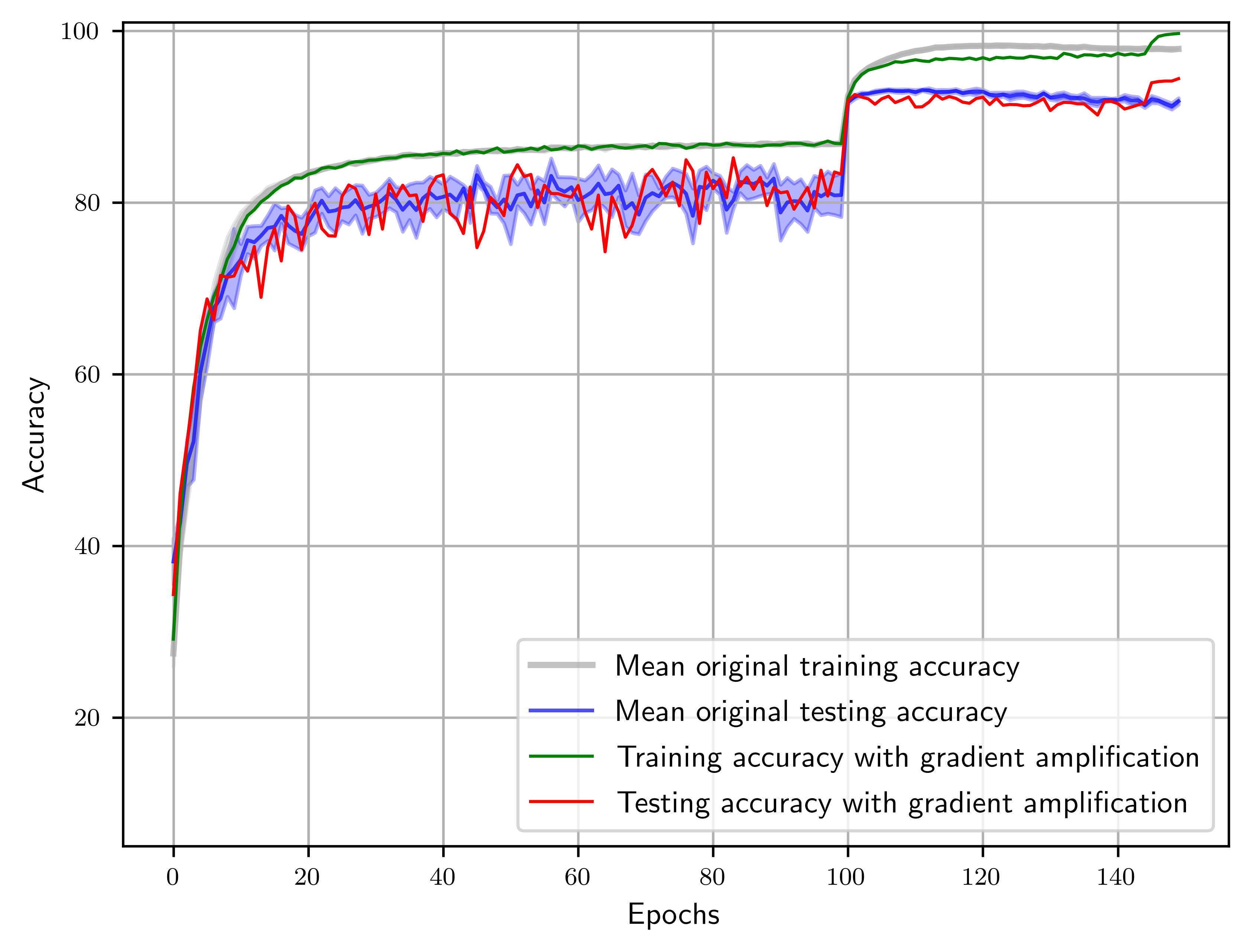}}
    \subfigure[Resnet-101 \label{fig:cifar10_formula_1_best_epoch-resnet-101}]{\includegraphics[width=0.2\textwidth]{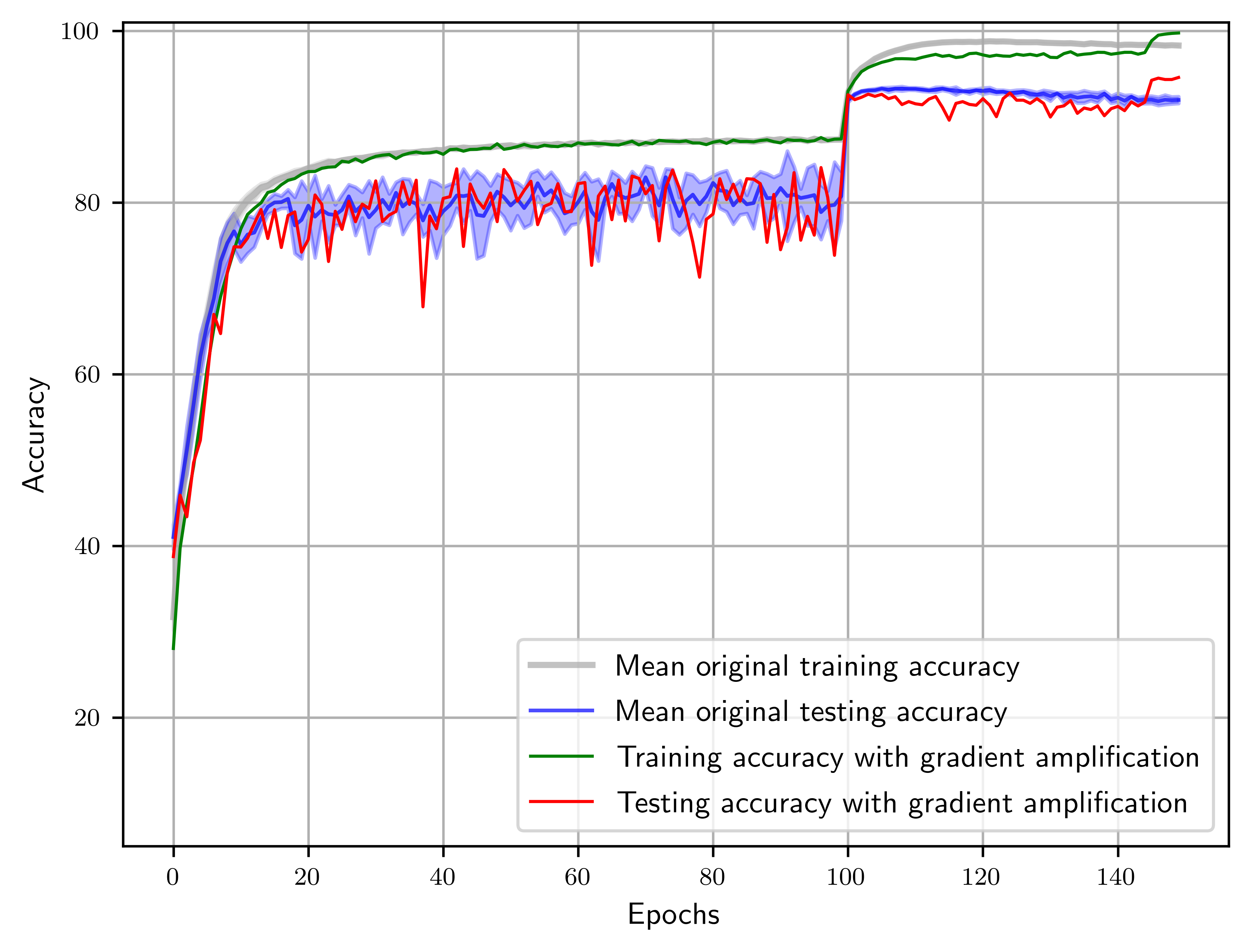}}
  \caption{Testing accuracies \% (Y-axis) of the best models on CIFAR-10 dataset with amplification performed using $G$ algorithm (\ref{alg:find_amp_formula1}) compared to original models without amplification. Mean training (gray) and mean testing (blue) accuracies (as well as original testing accuracies in light blue) are plotted, along with amplified training (green) and testing (red) accuracies.
 }
  \label{fig:form1_best_model_epochs_cifar10}
\end{figure*}

\definecolor{LightCyan}{rgb}{0.88,1,1}
\definecolor{Gray}{gray}{0.85}
\def\mathbi#1{\textbf{\em #1}}

\begin{table*}[h]
  \caption{Performance comparison of random, $G$, $G'$ layer-based gradient amplification models on CIFAR-10 dataset.  }
  \label{tab:best_models_intel_layer_amp_cifar10}
  \centering
  \resizebox{.99\textwidth}{!}{%

\renewcommand{\arraystretch}{1.4}

  \begin{tabular}{c l l l l l l}
    \hline
    & & \textit{Random layer}&\multicolumn{2}{c}{$Ours$ using $\hat{G}$} &\multicolumn{2}{c}{$Ours$ using $\hat{G'}$}\\
   \cline{4-7}
   \multirow{-2}{*}{Model} &\multirow{-2}{*}{$Original$ } &\multicolumn{1}{c}{\textit{amplification}\cite{basodi2020gradient} } & \multicolumn{1}{|c|} {Case-1} & \multicolumn{1}{|c|}{Case-2} & \multicolumn{1}{|c|} {Case-1} & \multicolumn{1}{|c|}{Case-2} \\
    \hline
    $VGG\_19$ &91.08\%  & 93.35 \% (\textbf{+2.27\%})& 93.30\% (\textbf{+2.22\%})& 92.92\% (\textbf{+1.84\%}) & 93.29\% (\textbf{+2.21\%})& 93.34\% (\textbf{2.26+\%})\\
    $Resnet\_18$ &92.39\% & 94.57\% (\textbf{+2.18\%}) & 93.90\% (\textbf{+1.51\%}) & 93.76\% (\textbf{+1.37\%})& 94.49\% (\textbf{+2.1\%})& 94.1\% (\textbf{+1.71\%}) \\
    $Resnet\_34$ &92.71\% & 94.39\%(\textbf{+1.68\%}) & 94.05\% (\textbf{+1.34\%}) & 93.89\% (\textbf{+1.18\%})& 94.56\% (\textbf{+1.85\%})& 94.14\% (\textbf{+1.43\%}) \\
    $Resnet\_50$ &91.80\%& 92.68\% (\textbf{+0.88\%}) &  94.24\% (\textbf{+2.44\%}) & 94.43\% (\textbf{+2.63\%}) & 94.34\% (\textbf{+2.54\%})& 94.02\% (\textbf{+2.22\%})\\
    $Resnet\_101$ &91.95\% & 93.04\% (\textbf{+1.09\%}) & 94.57\% (\textbf{+2.62\%}) & 94.54\% (\textbf{+2.59\%})& 94.35\% (\textbf{+2.4\%})& 94.7\% (\textbf{+2.75\%}) \\


\hline
    \hline
  \end{tabular}
  }
\end {table*}

\begin{table*}[h]
  \caption{Performance comparison of random, $G$, $G'$ layer-based gradient amplification models on CIFAR-100 dataset.  }
  \label{tab:best_models_intel_layer_amp_cifar100}
  \centering
  \resizebox{.99\textwidth}{!}{%

\renewcommand{\arraystretch}{1.4}
  \begin{tabular}{c l l l l l l}
    \hline
    & & \textit{Random layer} &\multicolumn{2}{c}{$Ours$ using $\hat{G}$} &\multicolumn{2}{c}{$Ours$ using $\hat{G'}$}\\
   \cline{4-7}
   \multirow{-2}{*}{Model} &\multirow{-2}{*}{$Original$  } &\multicolumn{1}{c}{\textit{amplification}\cite{basodi2020gradient} } &  \multicolumn{1}{|c|} {Case-1} & \multicolumn{1}{|c|}{Case-2} & \multicolumn{1}{|c|} {Case-1} & \multicolumn{1}{|c|}{Case-2} \\
    \hline
    $VGG\_19$ &65.27\%  & 66.52\% (\textbf{+1.25\%}) & 69.38\% (\textbf{+4.11\%})& 68.5\% (\textbf{+3.23\%}) & 69.83\% (\textbf{+4.56\%})& 69.25\% (\textbf{3.98+\%})\\
    $Resnet\_18$ &71.94\%  & 72.7\% (\textbf{+0.760\%}) & 75.33\% (\textbf{+3.39\%}) & 75.41\% (\textbf{+3.47\%})&76.14\% (\textbf{+4.2\%})& 75.35\% (\textbf{+3.41\%}) \\
    $Resnet\_34$ &72.18\%  &73.02\% (\textbf{+0.84\%})  & 74.86\% (\textbf{+2.68\%}) & 75.59\% (\textbf{+3.41\%})& 75.95\% (\textbf{+3.77\%})& 75.9\% (\textbf{+3.72\%}) \\
    $Resnet\_50$ &72.32\%  &73.05\% (\textbf{+0.73\%})  &  77.21\% (\textbf{+4.89\%}) & 77.43\% (\textbf{+5.11\%}) & 76.89\% (\textbf{+4.57\%})& 76.97\% (\textbf{+4.65\%})\\
    $Resnet\_101$ &73.00\%  &73.72\% (\textbf{+0.72\%})  & 77.63\% (\textbf{+4.63\%}) & 77.51\% (\textbf{+4.51\%})& 77.53\% (\textbf{+4.53\%})& 77.63\% (\textbf{+4.63\%}) \\

\hline
    \hline
  \end{tabular}
  }
\end {table*}

\subsection{Best models}
Here, we compare the best results of amplified models in each case with their corresponding original models without amplification. Testing accuracies of the best amplified models are shown in the table \ref{tab:best_models_intel_layer_amp_cifar10} and \ref{tab:best_models_intel_layer_amp_cifar100} for CIFAR-10 and CIFAR-100 datasets. Training and testing accuracies for each epoch of these best models with amplification along with the original models. Fig. \ref{fig:form1_best_model_epochs_cifar10} shows the best performing models while using measure $G$ for CIFAR-10 dataset. Similar improvements are observed for while using $G'$ on CIFAR-10 and also while using these measures on CIFAR-100 dataset. Since the mean accuracy of the original models are compared, their training accuracies(in gray), testing accuracies (in blue) including their mean accuracies are plotted along with amplified training(in green) and testing(in red) accuracies. These plots signify the importance of having the final epochs of the model to be trained without amplification and also demonstrate that the models do not overfit while training with amplification.

We also perform random amplification for deeper resnet models using some of the hyperparameters which have the better performance for resnet-18, resnet-34 models. The best accuracies of these models are also compared in the table below for both CIFAR-10 and CIFAR-100 datasets. Our results with amplification based on $G$ and $G'$ have similar performance and sometimes improved for VGG-19, resnet-18 and resnet-34 models on CIFAR-10 dataset. Resnet-50 and resnet-101 more than 2\% improvement than original as well as randomly amplified models. In the case of  CIFAR-100 dataset, all the models based on $G$ and $G'$ have significant performance improvement compared to original and random amplification.

\section{Conclusion} \label{sec:conclusion}
We propose two measures to compute effective gradient direction of a layer during the training process. These measures are used to determine the amplification layers based on two amplification thresholding strategies. Detailed experiments are performed to analyze each of the measures and their amplification strategies for a range of thresholds. Experiments performed on CIFAR-10 and CIFAR-100 datasets using different models show that intelligent amplification gets better accuracy compared to original models without amplification, even when trained with higher learning rates. In this work, gradient amplification is experimented on VGG and resnet models. As this method can be easily integrated into various deep learning architectures\cite{liu2017survey}, future work could extend this method to deep belief networks\cite{le2008representational}, recurrent neural networks\cite{pearlmutter1995gradient}, attention networks\cite{chaudhari2019attentive}, fuzzy/hybrid neural networks \cite{liu2005adaptive, zhang2014comparison, mudiyanselage2019deep, zhang2014parallel, zhang2016efficient,zhou2014fuzzy} and graph neural networks\cite{wu2020comprehensive}.


\section*{CRediT authorship contribution statement}
\textbf{Sunitha Basodi:} Conceptualization, Methodology, Software, Validation, Investigation, Writing – original draft, Visualization. \textbf{Krishna Pusuluri:} Conceptualization, Methodology, Software, Investigation, Validation, Writing – review \& editing. \textbf{Xueli Xiao:} Methodology, Resources, Software, Writing – review \& editing. \textbf{Yi Pan:}  Supervision, Funding acquisition, Project administration, Conceptualization, Methodology, Investigation, Validation, Writing – review \& editing.

\section*{Declaration of competing interest}
All the authors declare that they have no competing interests.

\section*{Data availability}
Openly available datasets are used in this work.

\section*{Acknowledgments}
Research Computing Technology and Innovation Core (ARCTIC) resources, which is supported by the National Science Foundation Major Research Instrumentation (MRI) grant number CNS-1920024. We gratefully acknowledge the support of NVIDIA Corporation with the donation of the Tesla K40 GPU also used for this research.

\def\UrlBreaks{\do\/\do-}
\bibliographystyle{elsarticle-num}
\bibliography{project}








\end{document}